\documentclass{article}

\PassOptionsToPackage{numbers, compress}{natbib}

    \usepackage[final]{neurips_2025}

\usepackage[utf8]{inputenc} 
\usepackage[T1]{fontenc}    
\usepackage{hyperref}       
\usepackage{url}            
\usepackage{booktabs}       
\usepackage{amsfonts}       
\usepackage{nicefrac}       
\usepackage{microtype}      
\usepackage{xcolor}         

\usepackage{multirow}
\usepackage{adjustbox}
\usepackage{array}
\usepackage{wrapfig,lipsum}
\usepackage{enumitem}
\usepackage{caption}
\usepackage{hhline}
\usepackage{colortbl}
\usepackage[table,xcdraw]{xcolor}

\newcommand{\eg}{\textit{e}.\textit{g}.}
\usepackage{booktabs}
\usepackage{subcaption}
\usepackage{pifont}
\newcommand{\cmark}{\ding{51}}%
 \usepackage{amsmath}

\title{VR-Drive: Viewpoint-Robust End-to-End Driving with Feed-Forward 3D Gaussian Splatting}

\author{%
  Hoonhee Cho\textsuperscript{1}\thanks{Equal contribution}\quad
  Jae-Young Kang\textsuperscript{1}\footnotemark[1]\quad
  Giwon Lee\textsuperscript{1}\footnotemark[1]\quad
  Hyemin Yang\textsuperscript{1}\footnotemark[1]
  \\
  \textbf{Heejun Park\textsuperscript{1}\quad
  Seokwoo Jung\textsuperscript{2}\quad
  Kuk-Jin Yoon\textsuperscript{1}} 
  \\
  \textsuperscript{1}\,KAIST \quad \textsuperscript{2}\,42dot 
  \\
  [3pt]
\href{https://vrdriveneurips.github.io/}{\color{magenta} https://vrdriveneurips.github.io/}\\
}

\begin{document}

\maketitle

\begin{abstract}
End-to-end autonomous driving (E2E-AD) has emerged as a promising paradigm that unifies perception, prediction, and planning into a holistic, data-driven framework. However, achieving robustness to varying camera viewpoints, a common real-world challenge due to diverse vehicle configurations, remains an open problem. In this work, we propose VR-Drive, a novel E2E-AD framework that addresses viewpoint generalization by jointly learning 3D scene reconstruction as an auxiliary task to enable planning-aware view synthesis. Unlike prior scene-specific synthesis approaches, VR-Drive adopts a feed-forward inference strategy that supports online training-time augmentation from sparse views without additional annotations. To further improve viewpoint consistency, we introduce a viewpoint-mixed memory bank that facilitates temporal interaction across multiple viewpoints and a viewpoint-consistent distillation strategy that transfers knowledge from original to synthesized views. Trained in a fully end-to-end manner, VR-Drive effectively mitigates synthesis-induced noise and improves planning under viewpoint shifts. In addition, we release a new benchmark dataset to evaluate E2E-AD performance under novel camera viewpoints, enabling comprehensive analysis. Our results demonstrate that VR-Drive is a scalable and robust solution for the real-world deployment of end-to-end autonomous driving systems.
\end{abstract}

\section{Introduction}
The end-to-end autonomous driving (E2E-AD) system refers to the integration of all modules, including perception, prediction, and planning nodes. The end-to-end driving paradigm~\cite{E2E_DualAT_ICRA2024,E2E_imitation_learning_ICRA2018,E2E_implicit_affordance_2020_CVPR,E2E_Neat_CVPR2021,E2E_reasonnet_cvpr2023,E2E_survey_IEEE2024,E2E_prompting_ICRA2024,E2E_OccNet_ICCV2023,E2E_PlanKD_CVPR2024,E2E_occworld} has consistently gained attention as a holistic approach, wherein the perception and prediction tasks are effectively integrated to support planning. This integration enhances both performance and efficiency, favoring a unified model for the entire driving task. This data-driven approach, compared to traditional rule-based planning, is designed to function robustly in complex scenarios by integrating various perception tasks (\eg,~detection, tracking, mapping, \textit{etc.}). During the training process, it incorporates vast amounts of data and annotations to enhance its capabilities.

Despite significant advancements and strong performance across various scenarios, existing end-to-end autonomous driving (E2E-AD) must evolve into scalable and flexible holistic models to become viable industry solutions. Recent E2E-AD systems~\cite{E2E_DualAD_CVPR2024,E2E_BridgeAD_CVPR2025,E2E_goalflow_CVPR2025,E2E_E2EAD_ICLR2025,E2E_Genad_ECCV2024,E2E_MomAD_CVPR2025,E2E_PPAD_ECCV2024,E2E_LAW_ICLR2025}, in particular, aim to achieve comparable performance using only raw camera input. However, the viewpoint of the camera~\cite{nvidia-sensors_config, philion2020lift} can vary depending on the vehicle's type and make, and systems that can effectively adapt to these changes are crucial from a real-world application perspective. A straightforward solution to this challenge would be to collect data using a variety of vehicles and camera rigs, and then use this data during the training process. However, this approach is impractical because it is impossible to pre-build camera viewpoints for every type of vehicle. Additionally, E2E-AD networks require annotations for various tasks, which incur significant costs, making it an impractical direction. Furthermore, to be deployable across different types of vehicles, the model must be flexible and robust not only to the predefined data but also to out-of-distribution (OOD) data. Therefore, the network must also ensure its generalization ability during the training process.

\begin{figure*}[t]
\begin{center}
\includegraphics[width=.99\linewidth]{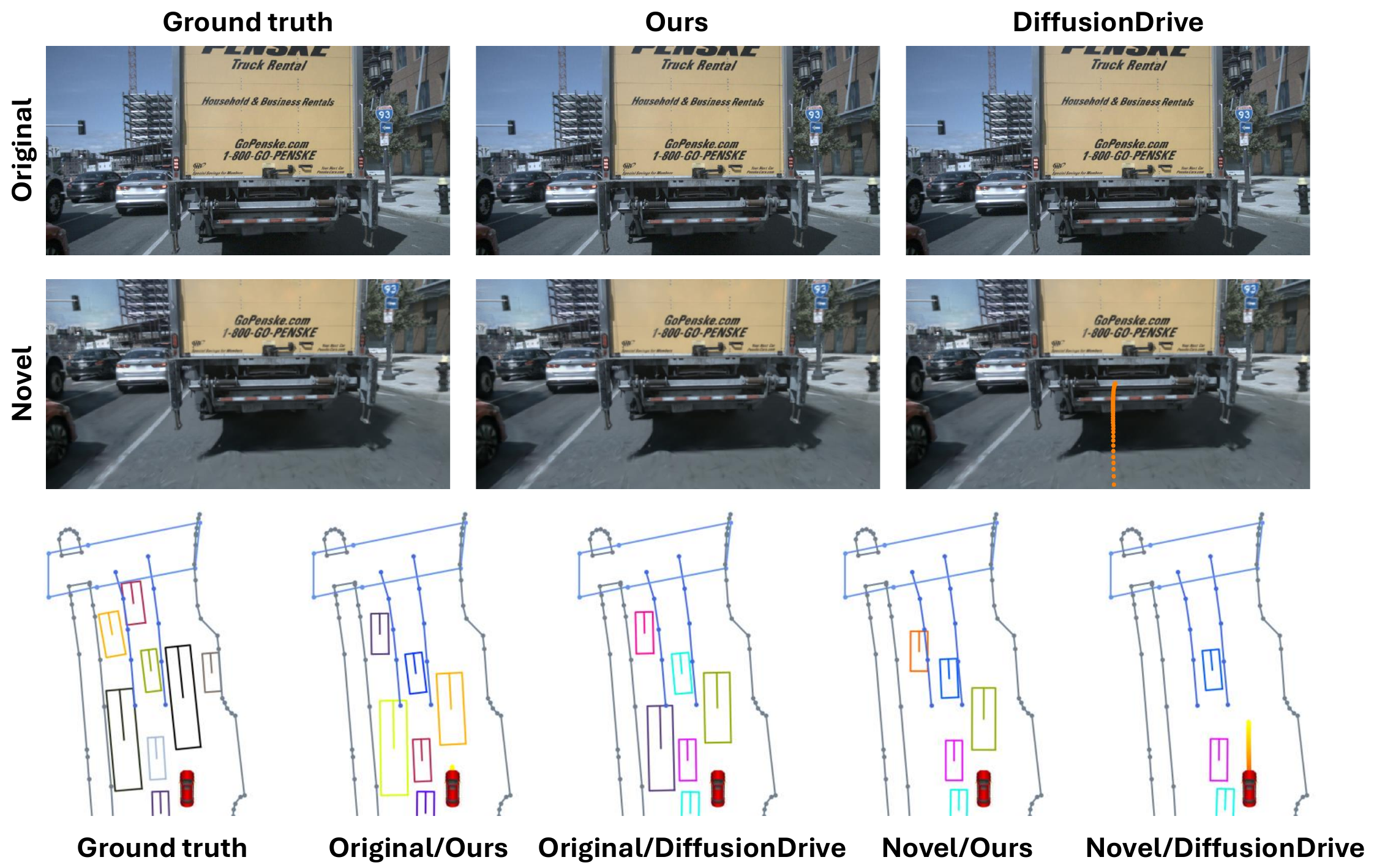}
\caption{\textbf{Example scenario where surrounding vehicles have stopped at a traffic signal.} In the original training view, both our VR-Drive and DiffusionDrive~\cite{E2E_Diffusiondrive} perform well in perceiving nearby vehicles and planning. However, with a lowered camera height, DiffusionDrive fails to detect surrounding vehicles, leading to a trajectory that collides with the front vehicle, posing a safety risk. In contrast, VR-Drive maintains accurate perception (except for those occluded due to the lowered camera height) and plans trajectories as effectively as in the original view.
}
\label{fig:teaser}
\end{center}
\vspace{-12pt}
\end{figure*}

To this end, we tackle the critical real-world challenge of generalization to diverse camera rigs in end-to-end autonomous driving (E2E-AD) systems. Specifically, we propose VR-Drive, which jointly learns 3D scene reconstruction as an auxiliary modular task within E2E-AD to augment the diversity of camera viewpoints. While numerous prior works~\cite{huang2024textit, mildenhall2021nerf, zhou2024drivinggaussian, khan2024autosplat} have explored novel view synthesis through 3D reconstruction, these methods are typically scene-optimized and require significant computational resources, making them unsuitable for real-time downstream tasks. Therefore, we advocate for an online scene reconstruction approach that operates effectively with sparse views. To this end, we adopt a feed-forward inference strategy~\cite{tian2025drivingforward, wang2024distillnerf, chen2024mvsplat, charatan2024pixelsplat} to ensure efficiency. Rather than training a separate novel view synthesis model, we integrate it as a joint modular task within the end-to-end framework, thereby reducing training complexity. Moreover, to prevent errors in view synthesis from propagating and degrading the final planning performance, VR-Drive introduces a unified framework that incorporates 3D reconstruction as an auxiliary task within E2E-AD, enabling novel view synthesis without requiring additional annotations. To learn a viewpoint-robust and consistent feature space, VR-Drive utilizes a viewpoint-mixed memory bank that encourages interaction between features from different viewpoints in the sequential training process by allowing them to mix in 3D space. Additionally, to mitigate the potential noise embedded in the features extracted from viewpoint-augmented images, we propose a distillation strategy that transfers knowledge from the original view features to guide the learning of these synthesized features. Benefiting from its end-to-end joint training, this planning-aware synthesis strategy ensures that the model remains effective under viewpoint shifts and contributes to improved downstream planning. 
As shown in Fig.~\ref{fig:teaser}, VR-Drive maintains robust performance under varying camera viewpoints, unlike existing E2E-AD methods that are sensitive to such changes, demonstrating its potential as a scalable and reliable end-to-end autonomous driving solution for real-world deployment.




The main contributions and unique aspects of our work are summarized as follows:
\vspace{-5pt}
\begin{itemize}[itemsep=-2pt, leftmargin=*]
\item We tackle viewpoint robustness in end-to-end autonomous driving (E2E-AD) by jointly learning 3D reconstruction for planning-aware view synthesis, enabling training data augmentation across diverse viewpoints and improving generalization to unseen camera configurations.

\item We propose a viewpoint-mixed memory bank that enables temporal interaction between features from different viewpoints, and introduce a viewpoint-consistent distillation strategy that transfers knowledge from original viewpoint images to their corresponding augmented novel view synthesis images in a 3D space.

\item We introduce a new benchmark dataset for E2E-AD to evaluate robustness under novel camera viewpoints unseen during training.

\end{itemize}

\section{Related Works}
\subsection{End-to-End Autonomous Driving}


End-to-end autonomous driving (E2E-AD) aims to generate final driving plans directly from raw sensor inputs within an integrated framework, in contrast to conventional methods that separately train perception, prediction, and planning modules.
Previous E2E-AD works can be largely categorized into two major directions: (1) focusing on architecture and task exploration, and (2) leveraging high-level information distillation.
Architecture-based approaches, such as \cite{E2E_UniAD, E2E_Transfuser++, zhang2024graphad}, demonstrate that submodules within an integrated framework can be optimized to enhance the final planning performance.
The following works~\cite{E2E_VAD, E2E_BEVPlanner} further improved planning accuracy by removing certain auxiliary tasks, such as occupancy prediction and motion prediction. In contrast, \cite{E2E_paradrive} reorganized traditionally sequential auxiliary tasks into a parallel structure, while \cite{E2E_BEVPlanner} proposed a task-aware training strategy to better leverage task relationships in parallel settings.


Architecture-based methods rely on large-scale annotated data, but often struggle in diverse scenarios due to biased training distributions, leading to issues such as causal confusion and long-tail errors.
To address this, several studies have explored distilling actions and feature information from rule-based or reinforcement learning (RL)-based experts trained in privileged settings~\cite{E2E_TCP,E2E_roach,E2E_thinktwice,E2E_driveadapter}.
Additionally, there has been research on utilizing language models for scene representation, prediction, and planning, enhancing situational understanding and adaptability through the general knowledge embedded in large-scale foundation models~\cite{E2E_lmdrive,E2E_drivelm,E2E_Gpt-driver,E2E_vlp,E2E_vadv2,E2E_Vlm-ad}.

Despite various research directions in E2E-AD, no prior work has addressed the development of model architectures that are robust to novel sensor viewpoints. This challenge is particularly critical, as sensor viewpoint variation is an inevitable and realistic factor in real-world deployments, arising from differences in vehicle types, sensor configurations, and mounting positions. However, it remains difficult to address within existing E2E-AD architectures and training paradigms, which are heavily dependent on the sensor inputs seen during training. In this work, we take the first step toward overcoming this limitation by proposing a method that enhances robustness to unseen sensor views.

\subsection{Viewpoint-Robust Representations and Scene Reconstruction}

Early studies~\cite{madan2021and, madan2021small, coors2019nova, do2020surface} have shown that neural networks are vulnerable to viewpoint changes, especially under distribution shifts. 
While these studies explored adversarial viewpoints in 2D perception, more recent efforts~\cite{nvidia-sensors_config, gao2024magicdrive3d, chang2024unified} have extended this line of research to address viewpoint robustness in 3D perception tasks. They typically leverage novel view synthesis to generate images under varying camera viewpoints, aiming to train perception algorithms that are robust across diverse views. Research on novel view synthesis via 3D scene reconstruction~\cite{yan2024street, zhou2024drivinggaussian, wu20244d, lin2024vastgaussian, turki2024hybridnerf} has advanced significantly, particularly with the emergence of Neural Radiance Fields (NeRF)~\cite{mildenhall2021nerf} and 3D Gaussian Splatting (3DGS)~\cite{kerbl20233d}. However, most methods are scene-specific and require long training times, as they rely on scene-by-scene optimization.

To be applicable to scalable E2E-AD, a view augmentation strategy must satisfy two key requirements.
(1) Since the test-time camera viewpoint is not fixed and can vary widely, the model must be robust to arbitrary views. This requires synthesizing diverse novel views during training, which in turn demands real-time online processing for both training and inference.
(2) To be effective in driving scenes, the method must support 3D reconstruction even with sparse or low-overlap observations.
To meet these requirements, we adopt a feed-forward 3D gaussian splatting~\cite{szymanowicz2024splatter, zheng2024gps, charatan2024pixelsplat, chen2024mvsplat, tian2025drivingforward} that is both generalizable and capable of online training and inference. By incorporating 3D scene reconstruction as a sub-task within E2E-AD, we enhance scene-level understanding and achieve performance gains even for the original viewpoints. Furthermore, by jointly training the view synthesis and driving tasks in an end-to-end manner, we account for potential synthesis errors and demonstrate the feasibility of extending novel view synthesis as a practical means to improve viewpoint robustness in E2E-AD.

\section{Methods}

\subsection{Overall Framework}

\label{sec:overall}

\begin{figure*}[t]
\begin{center}
\includegraphics[width=.99\linewidth]{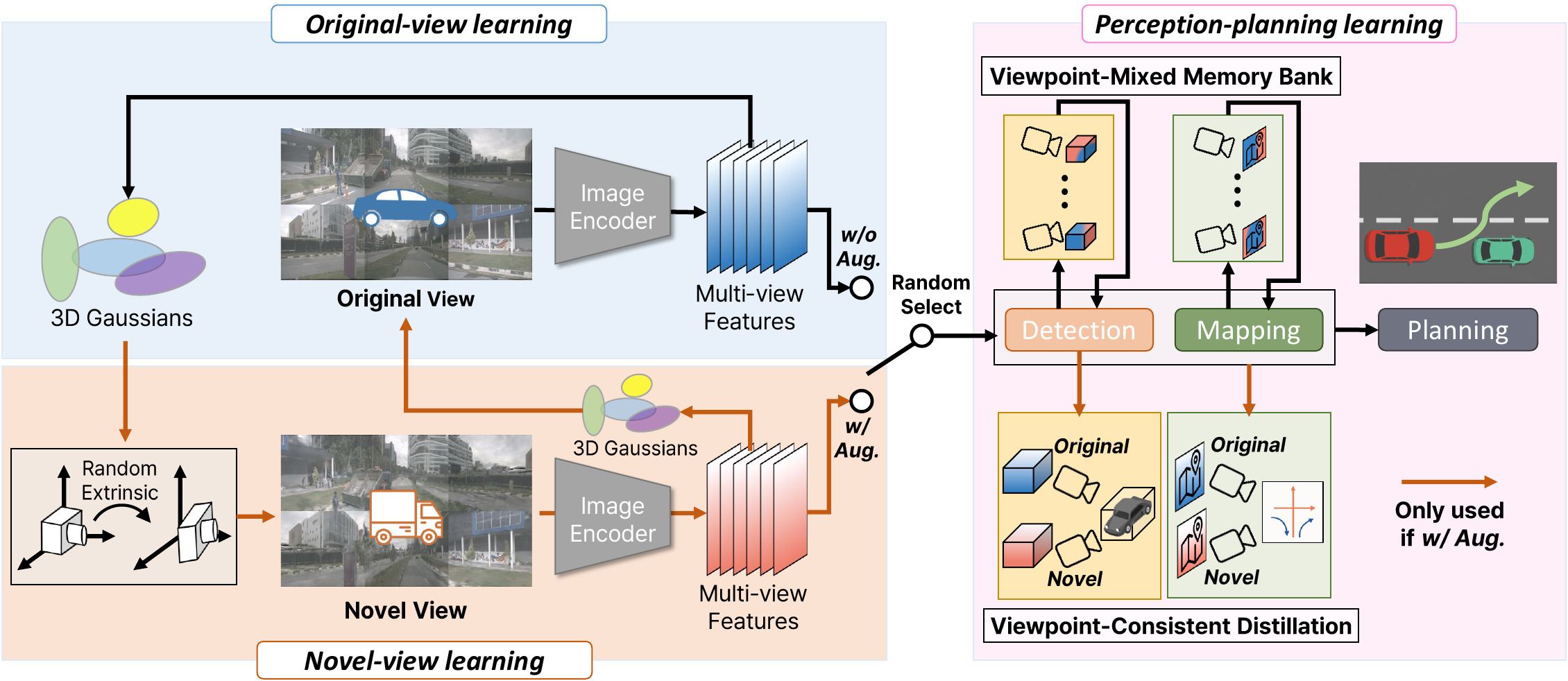}
\vspace{-3pt}
\caption{Overall framework of VR-Drive. Our overall framework consists of three main components, as follows: (1) original-view learning, (2) novel-view learning, and (3) perception-planning learning. For novel-view learning, the perception-planning head is randomly assigned to either the original or a novel view during training, allowing the model to generalize across different viewpoints.
}
\label{fig:overall}
\end{center}
\vspace{-7pt}
\end{figure*}

Given multi-view images, end-to-end autonomous driving (E2E-AD) models jointly learn perception and motion prediction to produce accurate motion plans for the ego vehicle. In addition to the standard pipeline of existing E2E-AD approaches, the proposed VR-Drive incorporates scene reconstruction as an auxiliary task, leveraging 3D Gaussian Splatting (3DGS)~\cite{kerbl20233d}. The overall framework of VR-Drive is shown in Fig.~\ref{fig:overall}. 
VR-Drive comprises three components, each targeting a distinct objective: (1) original-view learning, (2) novel-view learning, and (3) perception-planning learning.

\textbf{Original-view learning:}
During training, we use the original view as the default input of the pipeline. Given multi-view images, the image encoder (ResNet50~\cite{resnet}) first extracts original multi-view feature maps, $I\in\mathbb{R}^{N \times C \times H \times W}$, where $N$ is the number of camera views. These generated feature maps are utilized not only for perception and planning in autonomous driving, but also for learning and rendering novel views via 3DGS.
We build on the original 3DGS framework~\cite{kerbl20233d}, which represents a scene using Gaussian primitives ${g = (\mu, \Sigma, \alpha, c)}$, defined by position $\mu$, covariance $\Sigma$, opacity $\alpha$, and spherical harmonics for color $c$. The covariance matrix $\Sigma$ is constructed by combining the scaling factor $s$ and rotation quaternion $r$. Unlike the original 3DGS that relies on structure-from-motion for optimizing $\mu$, we predict primitives in a feed-forward~\cite{tian2025drivingforward}, pixel-wise manner directly from input images. Similar to previous work~\cite{E2E_Sparsedrive} that treated depth estimation as an auxiliary task within E2E-AD, we jointly learn depth as part of the E2E-AD framework. The estimated depth $D$ is then used to infer the position of Gaussian primitives $\mu \in \mathbb{R}^3$. We use the predicted depth map $D$ and the image feature map $I$, as input to a Gaussian network composed of multiple convolutional layers. This network predicts the remaining parameters of each Gaussian primitive, including the scaling factor $s \in \mathbb{R}^3_+$, rotation quaternion $r \in \mathbb{R}^4$, opacity $\alpha \in [0, 1]$, and color $c \in \mathbb{R}^k$ represented by $k$-degree spherical harmonics. To ensure valid ranges, we apply softplus to $H_s$ and softmax to $H_\alpha$, enforcing $s \in \mathbb{R}^3_+$ and $\alpha \in [0,1]$.  The feed-forward design enables online inference on novel views and generalization to new inputs without scene-specific constraints.



\textbf{Novel-view learning:}
VR-Drive aims to achieve robust planning performance by generating consistent feature representations even for camera viewpoints that were not observed during training. Specifically, at test time, it seeks to replicate the feature space of the original view across diverse, unseen viewpoints. 
To this end, we randomly sample camera extrinsics and render multi-view feature maps from arbitrary perspectives using the Gaussian primitives generated from the original view.
Given the rendered multi-view images from a novel view, we generate novel view features, $\tilde{I}\in\mathbb{R}^{N \times C \times H \times W}$, using a shared image encoder with the original view. Since the novel view features may differ in distribution from the original, we guide the model to generate feature representations that closely align with those of the original view. We observe that feed-forward 3DGS facilitates scene-level 3D understanding, which proves beneficial even under novel viewpoints. To encourage robustness, we additionally employ a cyclic reconstruction loss that trains the model to regenerate the original view from a novel one.

\begin{figure*}[t]
\begin{center}
\includegraphics[width=.98\linewidth]{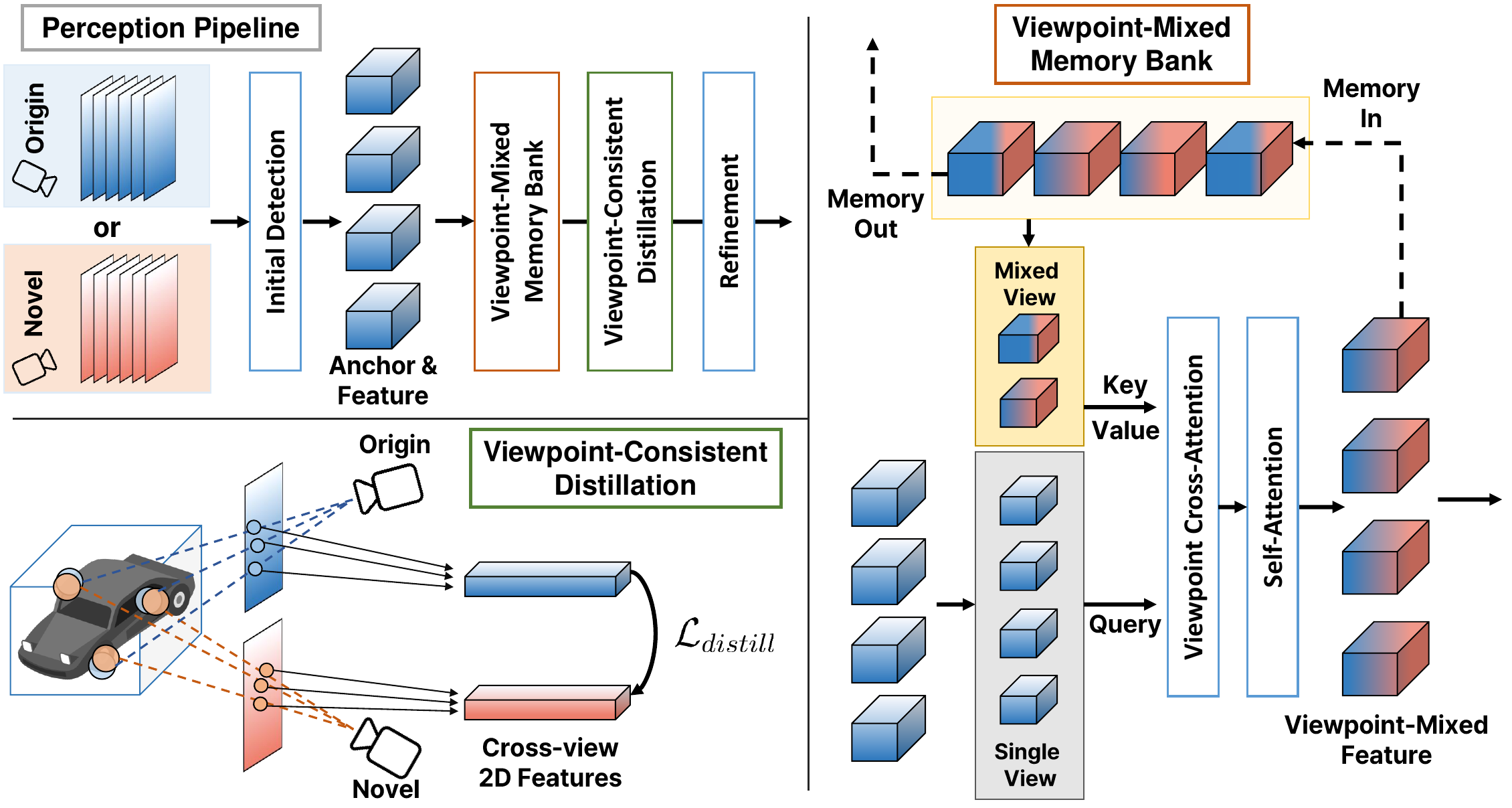}
\vspace{-4pt}
\caption{Illustration of the perception pipeline. VR-Drive includes two complementary techniques to ensure consistent feature representations across camera viewpoints: Viewpoint-Mixed Memory Bank and Viewpoint-Consistent Distillation.
}
\label{fig:module}
\end{center}
\vspace{-12pt}
\end{figure*}

\textbf{Perception-planning learning:}
VR-Drive selectively trains on original and novel views during the training to achieve robustness across diverse camera viewpoints. The image features extracted from the selected view are passed to the perception and planning heads, enabling planning based on the corresponding perception representation. Following~\cite{E2E_Diffusiondrive, E2E_drivetransformer, E2E_VAD}, we adopt 3D object detection and mapping as our perception tasks. 
More specifically, to achieve efficient representation, we utilize a same sparse architecture that leverages anchor- and instance feature-based designs~\cite{lin2022sparse4d, lin2023sparse4d} for both detection and mapping tasks. Since the two tasks differ only in the dimensionality of the anchors, we provide all descriptions and definitions in the context of detection, which are equally applicable to mapping. We first generate initial bounding box proposals using the detection module~\cite{lin2022sparse4d}, denoted as $B = \{B^1, B^2, \ldots, B^M\} \in \mathbb{R}^{M \times N_B}$, where $M$ is the number of anchors and $N_B$ is the dimensionality of each anchor. For each proposal, we also extract the corresponding instance features $F = \{ F^1, F^2, \ldots, F^M \} \in \mathbb{R}^{M \times N_i}$, where $N_i$ is the dimension of the instance feature. This allows us to encode the surrounding agents in the 3D space based on the extracted image features.
As illustrated in Fig.~\ref{fig:module}, we insert viewpoint-robust modules into the perception pipeline for detection and mapping, in addition to the conventional detection components. Specifically, we introduce two dedicated components within the perception stage of VR-Drive: the \textbf{Viewpoint-Mixed Memory Bank} and the \textbf{Viewpoint-Consistent Distillation} strategy, designed to address feature variations across viewpoints and promote canonical feature learning. We obtain the final perception results by refining the viewpoint-robust features through an additional detection decoder. Finally, to enable planning that interacts with the predicted agents, we adopt the motion planner proposed in~\cite{E2E_Diffusiondrive}.



\subsection{Viewpoint-Mixed Memory Bank}
As mentioned in Sec.~\ref{sec:overall}, the perception and planning pipeline randomly receives features from either the original view or a novel view during training. To promote canonical 3D feature learning from image inputs across diverse viewpoints with different distributions, we encourage interaction between 3D features extracted from diverse view during training. Rather than simply using a single pair of original and novel views, limiting the model to observing only two viewpoints within a single forward pass, we adopt a memory bank strategy that stores and updates features from continuously changing novel views to promote broader viewpoint generalization.  Let $F' \in \mathbb{R}^{M' \times N_i}$ be the instance features retrieved from the viewpoint-mixed memory bank, where $M'$ is the number of sampled features.
Following the method proposed in \cite{lin2022sparse4d}, we align $F'$ to the current frame by leveraging the velocities of the anchor box and the status of the ego vehicle to compensate for temporal shifts between viewpoints. Our objective is to generate interactive features between $F'$ and the instance features from the current view $F$. 
To achieve this, we leverage attention mechanisms~\cite{vaswani2017attention}
 to fuse features from the memory bank and the current view, resulting in the following mixed feature representation:
\begin{equation}
\mathbf{F} = \text{Cross-Attention}(Query = F, ~ Key = F',~ Value = F').
\label{eq:attention}
\end{equation}
The mixed feature, $\mathbf{F}$, is further processed through a self-attention mechanism to model interactions among agents, and are then passed to the viewpoint-consistent distillation module. The viewpoint-mixed memory bank is updated by selecting the top-$K$ high-confidence instances after the final refinement, while the oldest instances in the bank are discarded in an FIFO manner.



\subsection{Viewpoint-Consistent Distillation}

One potential challenge in learning viewpoint robustness through novel view synthesis is that the synthesized images may contain rendering artifacts, especially in occluded or texture-less regions. Moreover, novel view settings often involve more extreme or side-facing camera angles, which can be more challenging for autonomous driving due to reduced visibility or increased uncertainty in object localization. To address this, we adopt a distillation strategy in which the original view, typically containing more reliable and informative features due to better visibility and camera positioning, guides the learning of novel views. 
One simple strategy is to force alignment between two view features by projecting one onto the other using depth and pose. 
However, such alignment often excludes regions that are perceptually important for downstream tasks.
Instead, we utilize the instance features $\mathbf{F}$ and their corresponding anchor boxes $B$ to selectively distill information that is crucial from a planning perspective. Motivated by~\cite{wang2022detr3d}, we aim to extract representative object features by computing a learnable offset $\mathbf{p}$ and weight $\mathbf{w}$ for each instance $i$ based on its instance feature $\mathbf{F}_i$, defined as $\mathbf{p}_i = f(\mathbf{F_i}) \in \mathbb{R}^{s \times 3}$ and $\mathbf{w}_i = g(\mathbf{F_i}) \in \mathbb{R}^{N \times s}$, where $f$ and $g$ are a learnable keypoint and weight generations. 
Here, $s$ and $N$ denote the number of sampled points and cameras, respectively. 
Then, we compute the $j$-th 3D sampled point as follows:
\begin{equation}
\mathbf{p}_{i,j}^* = \mathbf{p}_{i,j} + \text{position}(B_i),
\label{eq:sampled_point}
\end{equation}
where $\text{position}(B_i)$ denotes the 3D center coordinates $(x, y, z)$ of the bounding box $B_i$. 
We project the sampled 3D points onto the image plane of each camera view using the corresponding transformation matrix, and extract image features at the $n$-th camera view via bilinear sampling, defined as:
\begin{equation}
\mathbf{f}_{n, i,j} = \text{BilinearSample}(I_n, \Pi_n \mathbf{p}^*_{i,j}) \in \mathbb{R}^{C}
\label{eq:bilinear_sample}
\end{equation}
where $\Pi_n$ is the camera transformation matrix and $I_n$ is the original view 2D image feature map from the $n$-th camera. Then, we define the aggregated feature at anchor index $i$ as:
\begin{equation}
S_i = \sum_n^N \sum_j^s \mathbf{w}_{n, i,j} \cdot \mathbf{f}_{n, i,j}.
\label{eq:weighted_aggregation}
\end{equation}
The same procedure is applied to the novel view image feature map $\tilde{I}$, resulting in $\tilde{S}$. To align the sampled features $\tilde{S}_i$ from the novel view with the corresponding features $S_i$ from the original view, we apply a mean squared error (MSE) loss between features. We restrict the loss of distillation to high-confidence anchors to avoid distillation in the background or noisy boxes. Let $\mathcal{I}^*$ be the set of anchors whose confidence scores exceed a predefined threshold $\tau$. The viewpoint-consistent distillation loss is defined as:
\begin{equation}
\mathcal{L}_{distill} = \frac{1}{|\mathcal{I}^*|} \sum_{i \in \mathcal{I}^*} \left\| \tilde{S}_i - \text{stopgrad}(S_i) \right\|_2^2,
\label{eq:filtered_distill_loss}
\end{equation}
where $\text{stopgrad}(\cdot)$ indicates gradient detachment.


Note that the viewpoint-mixed memory bank is always used, whereas the viewpoint-consistent distillation is only applied when a novel view image is used as the input for perception and planning.

\subsection{Loss Functions}
The loss functions consist of various tasks. For motion prediction and planning, we apply the winner-takes-all strategy~\cite{liang2020learning}.
In the planning task, an extra regression loss is introduced to handle ego status. For classification, we utilize focal loss~\cite{lin2017focal}, while L1 loss is used for regression in both detection and mapping tasks. Furthermore, L1 loss is also employed for depth estimation.
Additionally, we incorporate the viewpoint-consistent distillation loss. We also use a rendering loss for scene reconstruction, as described below.

\textbf{Rendering Loss.}
We use both L2 and LPIPS~\cite{zhang2018unreasonable} losses as the rendering objective. Since ground truth for various viewpoints is unavailable during training, we apply rendering loss through two alternative strategies, depending on whether novel view augmentation is used.
\vspace{-5pt}
\begin{itemize}[itemsep=-2pt, leftmargin=*]
\item[\textbf{-}] \textbf{Original Reconstruction Loss.}
The reconstruction loss encourages the model to render novel views from input images using Gaussian primitives. As real data lacks paired novel views, we simulate them by synthesizing adjacent-time views via splat-based rendering and apply the loss to the generated outputs.

\item[\textbf{-}] \textbf{Cyclic Reconstruction Loss.}
When a novel view is given as input for perception-planning heads, supervision using adjacent time-step images, as done with the original view, is not feasible due to the absence of paired frames. To support effective 3D scene learning with Gaussian primitives and depth, we adopt a cyclic rendering strategy that reconstructs the original view from the novel view.
\end{itemize}




The overall loss function for end-to-end training is:

\begin{equation}
\mathcal{L} = \mathcal{L}_{det} + \mathcal{L}_{map} + \mathcal{L}_{depth} + \mathcal{L}_{motion} + \mathcal{L}_{plan} + \mathcal{L}_{render} .
\label{eq:overall_loss}
\end{equation}


\begin{figure*}[t]
\begin{center}
\includegraphics[width=.99\linewidth]{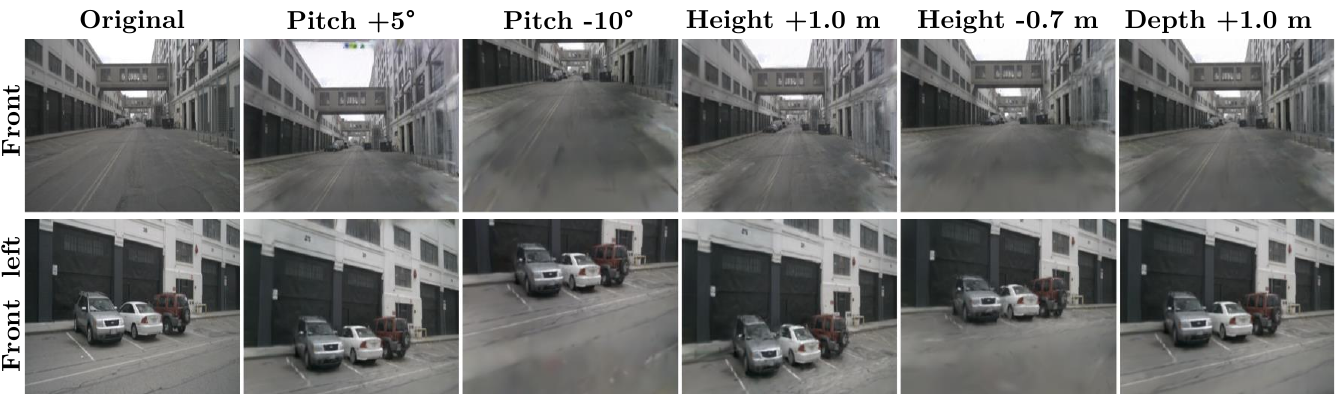}
\vspace{-4pt}
\caption{Variant camera viewpoints at test time, differing from the original training distribution.
}
\label{fig:data_sample}
\end{center}
\vspace{-20pt}
\end{figure*}

\section{End-to-End Autonomous Driving Benchmark with Viewpoint Variations}

\subsection{Training and Evaluation Setup}
Our work pioneers research on camera viewpoint variations in end-to-end autonomous driving (E2E-AD) and aims to establish a framework for training and evaluation in future studies. Considering the challenges of acquiring data with varying rigs during vehicle operation in real-world applications, we fix the rig to a single setup during the training process. Furthermore, our goal is to evaluate the model's robustness across various out-of-distribution data and assess its performance under different camera settings with distinct distributions. To achieve this, we introduce sensor variations at test time, deviating from the original camera configuration used during training, including: +5$^\circ$ pitch, -10$^\circ$ pitch, +1.0m height, -0.7m height, and +1.0m depth. These variations are configured based on the sensor settings from \cite{nvidia-sensors_config} to evaluate robustness.


\begin{table}[t]
\centering
\caption{Open-loop planning performance in nuScenes dataset. Metric calculation follows ST-P3~\cite{E2E_ST-P3}. The best performance in each setting is highlighted in \textbf{bold}. * denotes the usage of ego-status.}
\vspace{3pt}
\renewcommand{\arraystretch}{1.00}
\resizebox{1.0\textwidth}{!}{
\setlength\tabcolsep{7.0pt}
\begin{tabular}{c|c|c|ccc>{\columncolor{gray!20}}c|ccc>{\columncolor{gray!20}}c}
\specialrule{1pt}{0pt}{0pt}
\multicolumn{2}{c|}{\multirow{2.0}{*}{Camera Setting}} &\multirow{2.0}{*}{Methods} &  \multicolumn{4}{c|}{L2 (m) $\downarrow$} & \multicolumn{4}{c}{Collision Rate (\%) $\downarrow$}  \\
%
\cline{4-11}
\multicolumn{2}{c|}{} & & 1s & 2s & 3s &  \cellcolor{gray!20}Avg. & 1s & 2s & 3s &  Avg.
\\
\specialrule{1pt}{0pt}{0pt}
\multicolumn{2}{c|}{\multirow{6.0}{*}{Original}}
& AD-MLP*~\cite{E2E_ADMLP_archeive} & \textbf{0.20} & \textbf{0.26} &	\textbf{0.41} & \textbf{0.29} & 0.17 & 0.18 & 0.24 & 0.19\\
\multicolumn{2}{c|}{}& BEV-Planner*~\cite{E2E_BEVPlanner} & 0.28 &0.52 &	0.84 & 0.55 & 0.13 & 0.17 & 0.36 & 0.22 \\
\multicolumn{2}{c|}{}& VAD~\cite{E2E_VAD} & 0.41 & 0.70 & 1.05 & 0.72 & 0.07 & 0.17 & 0.41 & 0.22\\
\multicolumn{2}{c|}{}& SparseDrive~\cite{E2E_Sparsedrive} & 0.29 & 0.58 & 0.96 & 0.61 & \textbf{0.01} & 0.05 & 0.18 & 0.08 \\
\multicolumn{2}{c|}{}& DiffusionDrive~\cite{E2E_Diffusiondrive} & 0.27 & 0.54 & 0.90 & 0.57 & 0.03 & 0.05& 0.16 & 0.08\\
\multicolumn{2}{c|}{}& VR-Drive (Ours) & 0.29 & 0.57 & 0.95 & 0.60 & \textbf{0.01} & \textbf{0.03} & \textbf{0.14} & \textbf{0.06}\\
\specialrule{1pt}{0pt}{0pt}
\cellcolor{gray!20} & \multirow{6.0}{*}{pitch +5$^\circ$}

& AD-MLP*~\cite{E2E_ADMLP_archeive} & \textbf{0.20} & \textbf{0.26} &	\textbf{0.41} & \textbf{0.29} & 0.17 & 0.18 & 0.24 & 0.19\\
\cellcolor{gray!20}&& BEV-Planner*~\cite{E2E_BEVPlanner} & 0.29 & 0.56 & 0.91 & 0.59 & 0.27 & 0.31 & 0.54 & 0.37 \\

\cellcolor{gray!20}&& VAD~\cite{E2E_VAD} & 0.38	& 0.66 & 1.00 &	0.68 &	0.11 & 0.21 & 0.51 & 0.28\\
\cellcolor{gray!20}&& SparseDrive~\cite{E2E_Sparsedrive} & 0.32 & 0.63 & 1.03 & 0.66 & 0.02 & 0.08 & 0.35 & 0.15\\
\cellcolor{gray!20}&& DiffusionDrive~\cite{E2E_Diffusiondrive} & 0.33	& 0.64 & 1.04 & 0.67 & \textbf{0.00} & 0.09 & 0.24 & 0.11\\
\cellcolor{gray!20}&& VR-Drive (Ours) & 0.29 & 0.57 & 0.94 & 0.60 & \textbf{0.00} & \textbf{0.02} & \textbf{0.14} & \textbf{0.06} \\
\cline{2-11}
\cellcolor{gray!20} & \multirow{6.0}{*}{pitch -10$^\circ$} 
& AD-MLP*~\cite{E2E_ADMLP_archeive} & \textbf{0.20} & \textbf{0.26} &	\textbf{0.41} & \textbf{0.29} & 0.17 & 0.18 & \textbf{0.24} & 0.19\\
\cellcolor{gray!20}&& BEV-Planner*~\cite{E2E_BEVPlanner} & 0.27 & 0.51 & 0.86 & 0.54 & 0.64 & 0.73 & 0.93 & 0.76 \\

\cellcolor{gray!20}&&VAD~\cite{E2E_VAD} & 0.70 & 1.01 & 1.35 & 1.02 & 0.55 & 0.82 & 1.27 & 0.88\\
\cellcolor{gray!20}&& SparseDrive~\cite{E2E_Sparsedrive} & 0.46 & 0.91 & 1.50 & 0.96 & 0.03 & 0.15 & 0.50 & 0.23\\
\cellcolor{gray!20}&& DiffusionDrive~\cite{E2E_Diffusiondrive} & 0.45	& 0.91 & 1.52 & 0.96 & \textbf{0.02} & 0.16 & 0.55 & 0.24\\
\cellcolor{gray!20}&& VR-Drive (Ours) & 0.34 & 0.66 & 1.10 & 0.70 & \textbf{0.02} & \textbf{0.08} & \textbf{0.24} & \textbf{0.11} \\
\cline{2-11}
\cellcolor{gray!20} & \multirow{6.0}{*}{height +1.0 m}
& AD-MLP*~\cite{E2E_ADMLP_archeive} & \textbf{0.20} & \textbf{0.26} &	\textbf{0.41} & \textbf{0.29} & 0.17 & 0.18 & \textbf{0.24} & 0.19\\
\cellcolor{gray!20}&& BEV-Planner*~\cite{E2E_BEVPlanner} & 0.28 & 0.54 & 0.88 & 0.57 & 0.20 & 0.22 & 0.44 & 0.29 \\
\cellcolor{gray!20}&& VAD~\cite{E2E_VAD} & 0.41 & 0.70 & 1.07 & 0.73 & 0.14 & 0.45 & 0.80 & 0.47\\
\cellcolor{gray!20}&& SparseDrive~\cite{E2E_Sparsedrive} & 0.42 & 0.83 & 	1.36 & 0.87 & 0.10 & 0.45 &	1.08 &	0.54 \\
\cellcolor{gray!20}&& DiffusionDrive~\cite{E2E_Diffusiondrive} & 0.81	& 1.44 & 2.14 & 1.46 & 0.17 & 0.78 & 1.47 & 0.81 \\
\cellcolor{gray!20}&& VR-Drive (Ours) & 0.34 & 0.66 & 1.07 & 0.69 & \textbf{0.00} & \textbf{0.05} & 0.28 & \textbf{0.11} \\
\cline{2-11}
\cellcolor{gray!20} & \multirow{6.0}{*}{height -0.7 m}
& AD-MLP*~\cite{E2E_ADMLP_archeive} & \textbf{0.20} & \textbf{0.26} &	\textbf{0.41} & \textbf{0.29} & 0.17 & 0.18 & \textbf{0.24} & 0.19\\
\cellcolor{gray!20}&& BEV-Planner*~\cite{E2E_BEVPlanner} & 0.29 & 0.55 & 0.89 & 0.58 & 0.49 & 0.61 & 0.82 & 0.64 \\
\cellcolor{gray!20}&&
VAD~\cite{E2E_VAD} & 0.41 & 0.71 & 1.09 & 0.74 & 0.09 & 0.17 & 0.39 & 0.22\\
\cellcolor{gray!20}&& SparseDrive~\cite{E2E_Sparsedrive} & 0.50 & 0.97 & 1.56 & 1.01 & 0.01 & 0.20 & 0.68 & 0.30 \\
\cellcolor{gray!20}&& DiffusionDrive~\cite{E2E_Diffusiondrive} & 0.64 &	1.18 & 1.82 & 1.21 &  \textbf{0.00} & 0.12 & 0.49 & 0.20 \\
\cellcolor{gray!20}&& VR-Drive (Ours) & 0.34 & 0.66 & 1.09 & 0.69 & 0.03 & \textbf{0.11} & 0.28 & \textbf{0.14}  \\
\cline{2-11}
\cellcolor{gray!20} & \multirow{6.0}{*}{depth +1.0 m}
& AD-MLP*~\cite{E2E_ADMLP_archeive} & \textbf{0.20} & \textbf{0.26} &	\textbf{0.41} & \textbf{0.29} & 0.17 & 0.18 & \textbf{0.24} & 0.19\\
\cellcolor{gray!20}&& BEV-Planner*~\cite{E2E_BEVPlanner} & 0.29 & 0.55 & 0.89 & 0.58 & 0.17 & 0.23 & 0.43 & 0.28\\
\cellcolor{gray!20}&&
VAD~\cite{E2E_VAD} & 0.39 & 0.68 & 1.05 & 0.71 & 0.09 & 0.19 & 0.48 & 0.26 \\
\cellcolor{gray!20}&& SparseDrive~\cite{E2E_Sparsedrive} & 0.66 & 1.23 & 1.91 & 1.27 & 0.05 & 0.25 & 0.62 & 0.31\\
\cellcolor{gray!20}&& DiffusionDrive~\cite{E2E_Diffusiondrive} & 0.87	& 1.55 & 2.30 & 1.57 & 0.12 & 0.37 & 0.75 & 0.41 \\
\cellcolor{gray!20} && VR-Drive (Ours) & 0.37 & 0.69 & 1.11 & 0.72 & \textbf{0.02} & \textbf{0.11} & 0.27 & \textbf{0.13} \\
\cline{2-11}
\cellcolor{gray!20} & \multirow{6.0}{*}{Average}
& AD-MLP*~\cite{E2E_ADMLP_archeive} & \textbf{0.20} & \textbf{0.26} &	\textbf{0.41} & \textbf{0.29} & 0.17 & 0.18 & \textbf{0.24} & 0.19\\
\cellcolor{gray!20}&& BEV-Planner*~\cite{E2E_BEVPlanner} & 0.28 & 0.54 & 0.88 & 0.57 & 0.36 & 0.42 & 0.63 & 0.47\\
\cellcolor{gray!20}&&
VAD~\cite{E2E_VAD} & 0.46 & 0.75 & 1.11 & 0.78 & 0.20 & 0.37 & 0.69 & 0.42\\
\cellcolor{gray!20}&& SparseDrive~\cite{E2E_Sparsedrive} & 0.47 & 0.91 & 1.47 & 0.95 & 0.04 & 0.23 & 0.65 & 0.31\\
\cellcolor{gray!20}&& DiffusionDrive~\cite{E2E_Diffusiondrive} & 0.62 &  1.14 & 1.76 & 1.17 & 0.07 & 0.36 & 0.80 & 0.41\\
\cellcolor{gray!20} \multirow{-36.0}{*}{
Unseen}&& VR-Drive (Ours) & 0.34 & 0.65 & 1.06 & 0.68 & \textbf{0.01} & \textbf{0.07} & \textbf{0.24} & \textbf{0.11} \\
\specialrule{1pt}{0pt}{0pt}
\end{tabular}
}
\vspace{-15pt}
\label{tab:open_loop}
\end{table}

\subsection{Dataset Generation Protocol}
We use the nuScenes~\cite{Benchmarks_NuScenes} benchmark, which is widely used in recent E2E-AD works~\cite{lin2022sparse4d, E2E_drivetransformer, E2E_vadv2}. However, since the nuScenes dataset does not provide images from variant camera viewpoints, we performed offline scene optimization as a method to obtain data from various viewpoints. We performed offline scene optimization~\cite{chen2024omnire}, showcasing high-performance and strong geometric alignment, on nuScenes test sequences. This process enabled rendering various views for each sequence, as shown in Fig.~\ref{fig:data_sample}. 
After manually inspecting the test sequences, we excluded 4 sequences from the original 150 due to unsatisfactory quality, leaving 146 test sequences for unseen viewpoints. Note that this offline scene optimization requires significant training time for each scene, making it impractical for datasets with a large number of sequences. On an NVIDIA TITAN RTX, each sequence took over 8 hours to train, and the total time of optimization and rendering took more than 3 weeks. This underscores the practicality of our online novel view synthesis approach for training.








\section{Experiments}

\textbf{Experiment Setup.}
We evaluate the model using Average Displacement Error (ADE) and Collision Rate. For comparison, we use existing end-to-end models, including AD-MLP~\cite{E2E_ADMLP_archeive}, BEV-Planner~\cite{E2E_BEVPlanner}, VAD~\cite{E2E_VAD}, SparseDrive~\cite{E2E_Sparsedrive}, and DiffusionDrive~\cite{E2E_Diffusiondrive}.
During training, we rendered random novel view images with pitch in the range $[-10^{\circ}, 5^{\circ}]$, height in $ [-0.7\mathrm{m}, 1.0\mathrm{m}]$, and depth in $[-0.2\mathrm{m}, 1.0\mathrm{m}]$, which broadly covers the test configurations. Additional implementation details will be described in the supplementary material.

\textbf{Experimental Results.}
\label{sec:result}
Table~\ref{tab:open_loop} shows the performance of E2E-AD models on both original and novel views, where ``unseen'' refers to data that was not provided during training. When focusing on the performance in both the original and unseen domains, we begin by comparing the performance of our proposed VR-Drive with DiffusionDrive as an example. 
On the original domain, both models show similar performance. However, when evaluated on the unseen domain, DiffusionDrive experiences a significant increase in both ADE and collision rate. In contrast, our method demonstrates performance comparable to the original view, even in more challenging camera viewpoints under previously unseen distributions.




\begin{table*}[t]
\begin{center}
\centering
\setlength\tabcolsep{3.2pt}
\caption{\textbf{Ablation for design choices.} ``SR'' and ``VMM'' indicate scene reconstruction and viewpoint-mixed memory bank. ``CR'' and ``VCD'' indicate cyclic reconstruction loss and viewpoint-consistent distillation, respectively. $ \triangle $ indicates that scene reconstruction is learned jointly, but the generated novel view images are not used as perception and planning input.
}
\vspace{-2pt}
\resizebox{0.99\textwidth}{!}{
\begin{tabular}{c|cccc|cccc|cccc|cccc|cccc}
\toprule
& \multicolumn{4}{c|}{Modules} & \multicolumn{8}{c|}{Seen}  & \multicolumn{8}{c}{Unseen Average}  \\
\midrule
\multirow{2}{*}{ID} & \multirow{2}{*}{SR} & \multirow{2}{*}{VMM} & \multirow{2}{*}{CR} & \multirow{2}{*}{VCD} &   \multicolumn{4}{c|}{L2 (m) $\downarrow$} &\multicolumn{4}{c|}{Collision (\%) $\downarrow$} &   \multicolumn{4}{c|}{L2 (m) $\downarrow$} &\multicolumn{4}{c}{Collision (\%) $\downarrow$} \\
&  &  &  &  &  1s & 2s & 3s & \cellcolor{gray!30}Avg. & 1s & 2s & 3s & \cellcolor{gray!30}Avg. & 1s & 2s & 3s & \cellcolor{gray!30}Avg. & 1s & 2s & 3s & \cellcolor{gray!30}Avg. \\
\midrule
1 & - & - & - & - & 0.31 & 0.60 & 0.98 & \cellcolor{gray!20}0.63 & 0.13 & 0.10 & 0.19 &	\cellcolor{gray!20}0.14 & 0.47	&0.88&	1.38& \cellcolor{gray!20}0.91&	0.17&	0.25&	0.48& \cellcolor{gray!20}0.30 \\ 
2 & $ \triangle $ & - & - & - & \textbf{0.28}	&\textbf{0.56}&	\textbf{0.93}& \cellcolor{gray!20}\textbf{0.59}&	\textbf{0.00}&	0.05& 0.17&	\cellcolor{gray!20}0.07&0.46&	0.87&	1.36& \cellcolor{gray!20}0.90&	0.04&	0.20&	0.53& \cellcolor{gray!20}0.26\\ 
3 & \cmark & - & - & - & 0.31	&0.60	&0.97	& \cellcolor{gray!20}0.63	&0.03&	0.08&	0.20& \cellcolor{gray!20}0.10& 0.40&	0.76&	1.20& \cellcolor{gray!20}0.79&	0.04&	0.16&	0.36& \cellcolor{gray!20}0.19\\ 
4 & \cmark & \cmark & - & - & 0.31	&0.59&	0.95&	\cellcolor{gray!20}0.62&	0.02&	0.06&	0.19&	\cellcolor{gray!20}0.09 & 0.37&	0.70&	1.12&	\cellcolor{gray!20}0.73&	0.03&	0.13&	0.36& \cellcolor{gray!20}0.17 \\ 
5 & \cmark & \cmark & \cmark & - & 0.29&	\textbf{0.56} &	\textbf{0.93}&	\cellcolor{gray!20}\textbf{0.59}	&0.04&	0.06&	0.19&	\cellcolor{gray!20}0.09 & \textbf{0.33}	&\textbf{0.64}&	\textbf{1.06}&	\cellcolor{gray!20}\textbf{0.68}&	0.04&	0.13&	0.31&	\cellcolor{gray!20}0.16\\
6 & \cmark & \cmark & - & \cmark & 0.31&	0.59&	0.94&	\cellcolor{gray!20}0.61&	0.02&	0.05&	0.16&	\cellcolor{gray!20}0.08 & 0.37	&0.70	&1.14	&\cellcolor{gray!20}0.73	&0.02&	0.09&	0.31&	\cellcolor{gray!20}0.14\\
7 & \cmark & \cmark & \cmark & \cmark & 0.29 & 0.57 & 0.95 & \cellcolor{gray!20}0.60 & 0.01 & \textbf{0.03} & \textbf{0.14} & \cellcolor{gray!20}\textbf{0.06} & 0.34 & 0.65 & \textbf{1.06} & \cellcolor{gray!20}\textbf{0.68} & \textbf{0.01} & \textbf{0.07}& \textbf{0.24} & \cellcolor{gray!20}\textbf{0.11} \\
\bottomrule
\end{tabular}
\label{tab:design}
}
\end{center}
\vspace{-9pt}
\end{table*}

\begin{table*}[t]
\begin{center}
\centering
\setlength\tabcolsep{4.4pt}
\caption{Analysis of the range of random extrinsics for novel views during the training process.}
\vspace{-2pt}
\resizebox{0.99\textwidth}{!}{
\begin{tabular}{c|cccc|cccc|cccc|cccc}
\toprule
 & \multicolumn{8}{c|}{Seen}  & \multicolumn{8}{c}{Unseen Average}  \\
\midrule
 \multirow{2}{*}{Settings} &  \multicolumn{4}{c|}{L2 (m) $\downarrow$} &\multicolumn{4}{c|}{Collision (\%) $\downarrow$} &   \multicolumn{4}{c|}{L2 (m) $\downarrow$} &\multicolumn{4}{c}{Collision (\%) $\downarrow$} \\
&   1s & 2s & 3s & \cellcolor{gray!30}Avg. & 1s & 2s & 3s & \cellcolor{gray!30}Avg. & 1s & 2s & 3s & \cellcolor{gray!30}Avg. & 1s & 2s & 3s & \cellcolor{gray!30}Avg. \\
\midrule
 - &  \textbf{0.29} &\textbf{0.57} & \textbf{0.95} & \cellcolor{gray!20}\textbf{0.60} & 0.01 & \textbf{0.03} & \textbf{0.14} & \cellcolor{gray!20}\textbf{0.06} & 0.34 & \textbf{0.65} & \textbf{1.06} & \cellcolor{gray!20}\textbf{0.68} & \textbf{0.01} & \textbf{0.07} & \textbf{0.24} & \cellcolor{gray!20}\textbf{0.11} \\ 
 Superset & \textbf{0.29}	&\textbf{0.57}&	\textbf{0.95}&	\cellcolor{gray!20}\textbf{0.60}&	\textbf{0.00}&	\textbf{0.03}&	0.15&	\cellcolor{gray!20}\textbf{0.06} & \textbf{0.33}	&\textbf{0.65}	&\textbf{1.06}	&\cellcolor{gray!20}\textbf{0.68}	&0.03&	0.09&	0.25&	\cellcolor{gray!20}0.12\\ 
 Subset & 0.30& 0.60& 0.99& \cellcolor{gray!20}0.63 &\textbf{0.00} & 0.05& 0.16& \cellcolor{gray!20}0.07 &0.41	 &0.79	 &1.27 & \cellcolor{gray!20} 0.82& 0.02&  0.09& 0.27 &\cellcolor{gray!20}0.13\\

\bottomrule
\end{tabular}
\label{tab:view_range}
}
\end{center}
\vspace{-17pt}
\end{table*}


\section{Ablation Study}
\textbf{Effect of the components.}
We conducted an ablation study on each module, as shown in Table~\ref{tab:design}. Notably, comparing ID-1 and ID-2 reveals that simply enabling joint learning of scene reconstruction already improves performance on both original viewpoints. This suggests that online joint optimization with 3DGS contributes to improving the scalability of E2E-AD systems, likely by encouraging a more precise comprehension of 3D geometry. Such enhanced geometric understanding facilitates more informed and reliable planning decisions.
The most significant performance gain emerges at ID-3, where the novel view generated via scene reconstruction is used as an additional input to the model. Beyond this, the proposed modules further contribute to performance improvements. Interestingly, our method does not suffer from a trade-off where improved performance on novel views comes at the cost of degraded performance on original views. Instead, the proposed components enhance the model's overall capability, even in the original views. This suggests that novel views serve as an effective form of augmentation during training, and the introduced modules help guide the model to learn better representations, ultimately benefiting both original and novel view settings.



\textbf{Range of random extrinsics.}
We study the distribution shift between training and testing in terms of camera viewpoint diversity, as summarized in Table~\ref{tab:view_range}. 
For the experiments in Table~\ref{tab:open_loop}, we set the training-time random extrinsic ranges to pitch~$\in [-10^{\circ}, 5^{\circ}]$, height~$\in [-0.7\mathrm{m}, 1.0\mathrm{m}]$, and depth~$\in [-0.2\mathrm{m}, 1.0\mathrm{m}]$.
To examine generalization beyond the test distribution, the “Superset” setting expands the training sensor range to pitch$~\in[-15^{\circ}, 10^{\circ}]$, height$~\in [-1.0\mathrm{m}, 1.5\mathrm{m}]$, and depth$~  \in [-0.5\mathrm{m}, 1.5\mathrm{m}]$, 
covering viewpoints that go beyond the test distribution.
This allows us to investigate whether the model remains robust when trained with a broader range of viewpoints. Conversely, the “Subset” setting limits the sensor range to pitch$~\in[-5^{\circ}, 2^{\circ}]$, height$~\in [-0.3\mathrm{m}, 0.5\mathrm{m}]$, and depth$~  \in [-0.1\mathrm{m}, 0.5\mathrm{m}]$, ensuring that the training views do not overlap with any of the test-time configurations. 
Our model performs consistently across the Superset, Subset, and original settings, demonstrating robustness to continuous viewpoint variation.

\section{Closed-loop Evaluation on the CARLA dataset}

\begin{table}[h]
\vspace{-17pt}
\centering
\caption{Closed-loop test on CARLA dataset.}
\vspace{3pt}
\renewcommand{\arraystretch}{1.3}
\resizebox{1.0\textwidth}{!}{
\setlength\tabcolsep{2.4pt}
\begin{tabular}{l|cc||cc|cc|cc|cc|cc|cc}
\specialrule{1pt}{0pt}{0pt}
\multirow{3.0}{*}{Methods} &  \multicolumn{2}{c||}{\multirow{2.0}{*}{Original}} & \multicolumn{12}{c}{\cellcolor{gray!20}Unseen}  \\
\cline{4-15}
& & & \multicolumn{2}{c|}{pitch +5$^\circ$} & \multicolumn{2}{c|}{pitch -10$^\circ$} & \multicolumn{2}{c|}{height +1.0 m} & \multicolumn{2}{c|}{height -0.7 m} & \multicolumn{2}{c|}{depth +1.0 m} & \multicolumn{2}{c}{\textbf{Average}} \\
\cline{2-15}
& DS & RC & DS & RC & DS & RC & DS & RC & DS & RC & DS & RC & DS & RC
\\
\specialrule{1pt}{0pt}{0pt}
\multicolumn{15}{c}{Town05-\textit{Nov}}\\
\specialrule{1pt}{0pt}{0pt}
ST-P3~\cite{E2E_ST-P3} & 44.24&	100.00	&41.00&	100.00	&23.85&	100.00&	25.83&	100.00&	28.60	&100.00	&32.06&	100.00&	30.27&	100.00\\
TCP~\cite{E2E_TCP} & 92.73&	92.73&	70.33&	80.33&	4.65	&4.65&	88.51	&88.51&	0.00&	0.00	&91.11	&91.11&	50.92&	52.92\\
AD-MLP~\cite{E2E_ADMLP_archeive}& 13.59 &	32.83 &	13.59 &	32.83 &	13.59 & 32.83 &	13.59 & 32.83 & 13.59 &	32.83 &	13.59 & 32.83 &	13.59 &	32.83 \\
BEV-Planner~\cite{E2E_BEVPlanner} & 17.25 &	28.70 &	7.30 & 28.89 &	7.74 & 28.83 & 8.51 & 28.95 & 7.69 & 28.70 & 7.75 & 28.95 & 7.80 & 28.86 \\
Baseline & 76.47 & 99.20 & 69.41 & 89.60 & 45.65 & 99.38 & 48.67 & 100.00 & 41.59 & 86.76 & 35.95 & 98.60 & 48.25 & 94.87 \\
\textbf{VR-Drive (Ours)} & 84.04&	99.04&	75.00&	100.00&	91.26&	98.76&	98.44&	98.99&	80.67&	97.32&	95.88&	96.35&	88.25&	98.28\\
\specialrule{1pt}{0pt}{0pt}
\end{tabular}
}
\vspace{-6pt}
\label{tab:carla_result}
\end{table}

We use the CARLA 0.9.10.1 simulator~\cite{Benchmarks_Carla} for closed-loop testing.
For the closed-loop test, we evaluate performance using the Town05short benchmark.
We collect the training data from Town01, 02, 03, 04, 06, 07, and 10, using scenario routes based on previous work~\cite{E2E_Tranfuser}. 
For the evaluation, we assessed each model’s performance based on two key metrics: Driving Score (DS) and Route Completion (RC).
To provide a comprehensive comparison, we included several established end-to-end autonomous driving models. Specifically, we evaluated ST-P3\cite{E2E_ST-P3}, 
TCP\cite{E2E_TCP}, AD-MLP~\cite{E2E_ADMLP_archeive}, BEV-Planner~\cite{E2E_BEVPlanner}, and baseline 
alongside our proposed method. 
As the baseline, we adopt the ID-1 setting from Table~\ref{tab:design}, removing all proposed modules.

Following existing works~
\cite{E2E_PlanKD_CVPR2024, E2E_ST-P3, E2E_VAD, E2E_Tranfuser}, we adopt the Town05 benchmark for simulation. However, to enable training and evaluation on novel viewpoints, we establish a new benchmark. Specifically, we sample 20\% of sequences from Town05 Short to construct Town05-\textit{Nov}, which serves as our novel-view evaluation set.
For training data, we follow Transfuser~\cite{E2E_Tranfuser} and collect samples using the autopilot, but only from original viewpoints.
For fair comparison with prior works, we handle baselines based on their available resources. In the case of ST-P3~\cite{E2E_ST-P3} and TCP~\cite{E2E_TCP}, since pretrained checkpoints on Town05 are publicly released, we directly evaluate these models without retraining.

Table~\ref{tab:carla_result} shows the closed-loop evaluation results on the Town05-\textit{Nov} benchmark. Existing end-to-end autonomous driving approaches tend to struggle with planning in unseen test scenarios, sometimes failing to initiate driving in particularly challenging cases. Notably, the DS metric is more adversely affected compared to RC, experiencing degradation in perception performance when faced with novel viewpoint inputs. In contrast, our method demonstrates performance on unseen tests that is comparable to that of the original viewpoint.
\section{Conclusion}

In this work, we present VR-Drive, a unified end-to-end autonomous driving framework that leverages novel view synthesis and viewpoint-robust learning. To the best of our knowledge, we are the first to study camera viewpoint variation in E2E-AD for real-world applications. 
We benchmark VR-Drive on the nuScenes dataset and the CARLA simulator, achieving state-of-the-art performance across diverse camera viewpoints and out-of-distribution conditions.

\noindent
\textbf{Limitation and Future Work.} The performance of VR-Drive is influenced by the accuracy of camera calibration. While errors in calibration may lead to suboptimal results, the system could be made more robust to such errors. Addressing this issue and enhancing the system's robustness to calibration inaccuracies could be an important focus for future work.

\section*{Acknowledgement}
We thank 42dot for funding this research. We also appreciate the support and valuable discussions from the members of the 42dot research team. This work was supported by the National Research Foundation of Korea(NRF) grant funded by the Korea government(MSIT) (NRF2022R1A2B5B03002636).

\bibliographystyle{unsrtnat}
\bibliography{main}

\begin{thebibliography}{74}
\providecommand{\natexlab}[1]{#1}
\providecommand{\url}[1]{\texttt{#1}}
\expandafter\ifx\csname urlstyle\endcsname\relax
  \providecommand{\doi}[1]{doi: #1}\else
  \providecommand{\doi}{doi: \begingroup \urlstyle{rm}\Url}\fi

\bibitem[Chen et~al.(2024{\natexlab{a}})Chen, Yu, Li, You, and Tan]{E2E_DualAT_ICRA2024}
Zesong Chen, Ze~Yu, Jun Li, Linlin You, and Xiaojun Tan.
\newblock Dualat: Dual attention transformer for end-to-end autonomous driving.
\newblock In \emph{2024 IEEE International Conference on Robotics and Automation (ICRA)}, pages 16353--16359. IEEE, 2024{\natexlab{a}}.

\bibitem[Codevilla et~al.(2018)Codevilla, Müller, López, Koltun, and Dosovitskiy]{E2E_imitation_learning_ICRA2018}
Felipe Codevilla, Matthias Müller, Antonio López, Vladlen Koltun, and Alexey Dosovitskiy.
\newblock End-to-end driving via conditional imitation learning.
\newblock In \emph{2018 IEEE International Conference on Robotics and Automation (ICRA)}, pages 4693--4700, 2018.
\newblock \doi{10.1109/ICRA.2018.8460487}.

\bibitem[Toromanoff et~al.(2020)Toromanoff, Wirbel, and Moutarde]{E2E_implicit_affordance_2020_CVPR}
Marin Toromanoff, Emilie Wirbel, and Fabien Moutarde.
\newblock End-to-end model-free reinforcement learning for urban driving using implicit affordances.
\newblock In \emph{Proceedings of the IEEE/CVF Conference on Computer Vision and Pattern Recognition (CVPR)}, June 2020.

\bibitem[Chitta et~al.(2021)Chitta, Prakash, and Geiger]{E2E_Neat_CVPR2021}
Kashyap Chitta, Aditya Prakash, and Andreas Geiger.
\newblock Neat: Neural attention fields for end-to-end autonomous driving.
\newblock In \emph{Proceedings of the IEEE/CVF International Conference on Computer Vision}, pages 15793--15803, 2021.

\bibitem[Shao et~al.(2023)Shao, Wang, Chen, Waslander, Li, and Liu]{E2E_reasonnet_cvpr2023}
Hao Shao, Letian Wang, Ruobing Chen, Steven~L Waslander, Hongsheng Li, and Yu~Liu.
\newblock Reasonnet: End-to-end driving with temporal and global reasoning.
\newblock In \emph{Proceedings of the IEEE/CVF conference on computer vision and pattern recognition}, pages 13723--13733, 2023.

\bibitem[Chen et~al.(2024{\natexlab{b}})Chen, Wu, Chitta, Jaeger, Geiger, and Li]{E2E_survey_IEEE2024}
Li~Chen, Penghao Wu, Kashyap Chitta, Bernhard Jaeger, Andreas Geiger, and Hongyang Li.
\newblock End-to-end autonomous driving: Challenges and frontiers.
\newblock \emph{IEEE Transactions on Pattern Analysis and Machine Intelligence}, 2024{\natexlab{b}}.

\bibitem[Duan et~al.(2024)Duan, Zhang, and Xu]{E2E_prompting_ICRA2024}
Yiqun Duan, Qiang Zhang, and Renjing Xu.
\newblock Prompting multi-modal tokens to enhance end-to-end autonomous driving imitation learning with llms.
\newblock In \emph{2024 IEEE International Conference on Robotics and Automation (ICRA)}, pages 6798--6805. IEEE, 2024.

\bibitem[Tong et~al.(2023)Tong, Sima, Wang, Chen, Wu, Deng, Gu, Lu, Luo, Lin, et~al.]{E2E_OccNet_ICCV2023}
Wenwen Tong, Chonghao Sima, Tai Wang, Li~Chen, Silei Wu, Hanming Deng, Yi~Gu, Lewei Lu, Ping Luo, Dahua Lin, et~al.
\newblock Scene as occupancy.
\newblock In \emph{Proceedings of the IEEE/CVF International Conference on Computer Vision}, pages 8406--8415, 2023.

\bibitem[Feng et~al.(2024)Feng, Li, Ren, Yuan, and Wang]{E2E_PlanKD_CVPR2024}
Kaituo Feng, Changsheng Li, Dongchun Ren, Ye~Yuan, and Guoren Wang.
\newblock On the road to portability: Compressing end-to-end motion planner for autonomous driving.
\newblock In \emph{Proceedings of the IEEE/CVF Conference on Computer Vision and Pattern Recognition}, pages 15099--15108, 2024.

\bibitem[Zheng et~al.(2024{\natexlab{a}})Zheng, Chen, Huang, Zhang, Duan, and Lu]{E2E_occworld}
Wenzhao Zheng, Weiliang Chen, Yuanhui Huang, Borui Zhang, Yueqi Duan, and Jiwen Lu.
\newblock Occworld: Learning a 3d occupancy world model for autonomous driving.
\newblock In \emph{European conference on computer vision}, pages 55--72. Springer, 2024{\natexlab{a}}.

\bibitem[Doll et~al.(2024)Doll, Hanselmann, Schneider, Schulz, Cordts, Enzweiler, and Lensch]{E2E_DualAD_CVPR2024}
Simon Doll, Niklas Hanselmann, Lukas Schneider, Richard Schulz, Marius Cordts, Markus Enzweiler, and Hendrik Lensch.
\newblock Dualad: Disentangling the dynamic and static world for end-to-end driving.
\newblock In \emph{Proceedings of the IEEE/CVF Conference on Computer Vision and Pattern Recognition}, pages 14728--14737, 2024.

\bibitem[Zhang et~al.(2025)Zhang, Song, Jin, and Zhang]{E2E_BridgeAD_CVPR2025}
Bozhou Zhang, Nan Song, Xin Jin, and Li~Zhang.
\newblock Bridging past and future: End-to-end autonomous driving with historical prediction and planning.
\newblock \emph{arXiv preprint arXiv:2503.14182}, 2025.

\bibitem[Xing et~al.(2025)Xing, Zhang, Hu, Jiang, He, Zhang, Long, and Yin]{E2E_goalflow_CVPR2025}
Zebin Xing, Xingyu Zhang, Yang Hu, Bo~Jiang, Tong He, Qian Zhang, Xiaoxiao Long, and Wei Yin.
\newblock Goalflow: Goal-driven flow matching for multimodal trajectories generation in end-to-end autonomous driving.
\newblock \emph{arXiv preprint arXiv:2503.05689}, 2025.

\bibitem[Li and Cui()]{E2E_E2EAD_ICLR2025}
Peidong Li and Dixiao Cui.
\newblock Navigation-guided sparse scene representation for end-to-end autonomous driving.
\newblock In \emph{The Thirteenth International Conference on Learning Representations}.

\bibitem[Zheng et~al.(2024{\natexlab{b}})Zheng, Song, Guo, Zhang, and Chen]{E2E_Genad_ECCV2024}
Wenzhao Zheng, Ruiqi Song, Xianda Guo, Chenming Zhang, and Long Chen.
\newblock Genad: Generative end-to-end autonomous driving.
\newblock In \emph{European Conference on Computer Vision}, pages 87--104. Springer, 2024{\natexlab{b}}.

\bibitem[Song et~al.(2025)Song, Jia, Liu, Pan, Zhang, Wang, Zhang, Xu, Yang, and Luo]{E2E_MomAD_CVPR2025}
Ziying Song, Caiyan Jia, Lin Liu, Hongyu Pan, Yongchang Zhang, Junming Wang, Xingyu Zhang, Shaoqing Xu, Lei Yang, and Yadan Luo.
\newblock Don't shake the wheel: Momentum-aware planning in end-to-end autonomous driving.
\newblock \emph{arXiv preprint arXiv:2503.03125}, 2025.

\bibitem[Chen et~al.(2024{\natexlab{c}})Chen, Ye, Xu, Cao, and Chen]{E2E_PPAD_ECCV2024}
Zhili Chen, Maosheng Ye, Shuangjie Xu, Tongyi Cao, and Qifeng Chen.
\newblock Ppad: Iterative interactions of prediction and planning for end-to-end autonomous driving.
\newblock In \emph{European Conference on Computer Vision}, pages 239--256. Springer, 2024{\natexlab{c}}.

\bibitem[Li et~al.(2024{\natexlab{a}})Li, Fan, He, Wang, Chen, Zhang, and Tan]{E2E_LAW_ICLR2025}
Yingyan Li, Lue Fan, Jiawei He, Yuqi Wang, Yuntao Chen, Zhaoxiang Zhang, and Tieniu Tan.
\newblock Enhancing end-to-end autonomous driving with latent world model.
\newblock \emph{arXiv preprint arXiv:2406.08481}, 2024{\natexlab{a}}.

\bibitem[Klinghoffer et~al.(2023)Klinghoffer, Philion, Chen, Litany, Gojcic, Joo, Raskar, Fidler, and Alvarez]{nvidia-sensors_config}
Tzofi Klinghoffer, Jonah Philion, Wenzheng Chen, Or~Litany, Zan Gojcic, Jungseock Joo, Ramesh Raskar, Sanja Fidler, and Jose~M Alvarez.
\newblock Towards viewpoint robustness in bird's eye view segmentation.
\newblock In \emph{Proceedings of the IEEE/CVF International Conference on Computer Vision}, pages 8515--8524, 2023.

\bibitem[Philion and Fidler(2020)]{philion2020lift}
Jonah Philion and Sanja Fidler.
\newblock Lift, splat, shoot: Encoding images from arbitrary camera rigs by implicitly unprojecting to 3d.
\newblock In \emph{Computer Vision--ECCV 2020: 16th European Conference, Glasgow, UK, August 23--28, 2020, Proceedings, Part XIV 16}, pages 194--210. Springer, 2020.

\bibitem[Liao et~al.(2024)Liao, Chen, Yin, Jiang, Wang, Yan, Zhang, Li, Zhang, Zhang, et~al.]{E2E_Diffusiondrive}
Bencheng Liao, Shaoyu Chen, Haoran Yin, Bo~Jiang, Cheng Wang, Sixu Yan, Xinbang Zhang, Xiangyu Li, Ying Zhang, Qian Zhang, et~al.
\newblock Diffusiondrive: Truncated diffusion model for end-to-end autonomous driving.
\newblock \emph{arXiv preprint arXiv:2411.15139}, 2024.

\bibitem[Huang et~al.(2024)Huang, Wei, Zheng, An, Lu, Zhan, Tomizuka, Keutzer, and Zhang]{huang2024textit}
Nan Huang, Xiaobao Wei, Wenzhao Zheng, Pengju An, Ming Lu, Wei Zhan, Masayoshi Tomizuka, Kurt Keutzer, and Shanghang Zhang.
\newblock S3gaussian: Self-supervised street gaussians for autonomous driving.
\newblock \emph{arXiv preprint arXiv:2405.20323}, 2024.

\bibitem[Mildenhall et~al.(2021)Mildenhall, Srinivasan, Tancik, Barron, Ramamoorthi, and Ng]{mildenhall2021nerf}
Ben Mildenhall, Pratul~P Srinivasan, Matthew Tancik, Jonathan~T Barron, Ravi Ramamoorthi, and Ren Ng.
\newblock Nerf: Representing scenes as neural radiance fields for view synthesis.
\newblock \emph{Communications of the ACM}, 65\penalty0 (1):\penalty0 99--106, 2021.

\bibitem[Zhou et~al.(2024)Zhou, Lin, Shan, Wang, Sun, and Yang]{zhou2024drivinggaussian}
Xiaoyu Zhou, Zhiwei Lin, Xiaojun Shan, Yongtao Wang, Deqing Sun, and Ming-Hsuan Yang.
\newblock Drivinggaussian: Composite gaussian splatting for surrounding dynamic autonomous driving scenes.
\newblock In \emph{Proceedings of the IEEE/CVF conference on computer vision and pattern recognition}, pages 21634--21643, 2024.

\bibitem[Khan et~al.(2024)Khan, Fazlali, Sharma, Cao, Bai, Ren, and Liu]{khan2024autosplat}
Mustafa Khan, Hamidreza Fazlali, Dhruv Sharma, Tongtong Cao, Dongfeng Bai, Yuan Ren, and Bingbing Liu.
\newblock Autosplat: Constrained gaussian splatting for autonomous driving scene reconstruction.
\newblock \emph{arXiv preprint arXiv:2407.02598}, 2024.

\bibitem[Tian et~al.(2025)Tian, Tan, Xie, and Ma]{tian2025drivingforward}
Qijian Tian, Xin Tan, Yuan Xie, and Lizhuang Ma.
\newblock Drivingforward: Feed-forward 3d gaussian splatting for driving scene reconstruction from flexible surround-view input.
\newblock In \emph{Proceedings of the AAAI Conference on Artificial Intelligence}, volume~39, pages 7374--7382, 2025.

\bibitem[Wang et~al.(2024)Wang, Kim, Yang, Yu, Ivanovic, Waslander, Wang, Fidler, Pavone, and Karkus]{wang2024distillnerf}
Letian Wang, Seung~Wook Kim, Jiawei Yang, Cunjun Yu, Boris Ivanovic, Steven Waslander, Yue Wang, Sanja Fidler, Marco Pavone, and Peter Karkus.
\newblock Distillnerf: Perceiving 3d scenes from single-glance images by distilling neural fields and foundation model features.
\newblock \emph{Advances in Neural Information Processing Systems}, 37:\penalty0 62334--62361, 2024.

\bibitem[Chen et~al.(2024{\natexlab{d}})Chen, Xu, Zheng, Zhuang, Pollefeys, Geiger, Cham, and Cai]{chen2024mvsplat}
Yuedong Chen, Haofei Xu, Chuanxia Zheng, Bohan Zhuang, Marc Pollefeys, Andreas Geiger, Tat-Jen Cham, and Jianfei Cai.
\newblock Mvsplat: Efficient 3d gaussian splatting from sparse multi-view images.
\newblock In \emph{European Conference on Computer Vision}, pages 370--386. Springer, 2024{\natexlab{d}}.

\bibitem[Charatan et~al.(2024)Charatan, Li, Tagliasacchi, and Sitzmann]{charatan2024pixelsplat}
David Charatan, Sizhe~Lester Li, Andrea Tagliasacchi, and Vincent Sitzmann.
\newblock pixelsplat: 3d gaussian splats from image pairs for scalable generalizable 3d reconstruction.
\newblock In \emph{Proceedings of the IEEE/CVF conference on computer vision and pattern recognition}, pages 19457--19467, 2024.

\bibitem[Hu et~al.(2023)Hu, Yang, Chen, Li, Sima, Zhu, Chai, Du, Lin, Wang, et~al.]{E2E_UniAD}
Yihan Hu, Jiazhi Yang, Li~Chen, Keyu Li, Chonghao Sima, Xizhou Zhu, Siqi Chai, Senyao Du, Tianwei Lin, Wenhai Wang, et~al.
\newblock Planning-oriented autonomous driving.
\newblock In \emph{Proceedings of the IEEE/CVF conference on computer vision and pattern recognition}, pages 17853--17862, 2023.

\bibitem[Jaeger et~al.(2023)Jaeger, Chitta, and Geiger]{E2E_Transfuser++}
Bernhard Jaeger, Kashyap Chitta, and Andreas Geiger.
\newblock Hidden biases of end-to-end driving models.
\newblock In \emph{Proceedings of the IEEE/CVF International Conference on Computer Vision}, pages 8240--8249, 2023.

\bibitem[Zhang et~al.(2024)Zhang, Qian, Li, Pan, Chen, Liang, Zhang, Zhang, Li, Fu, et~al.]{zhang2024graphad}
Yunpeng Zhang, Deheng Qian, Ding Li, Yifeng Pan, Yong Chen, Zhenbao Liang, Zhiyao Zhang, Shurui Zhang, Hongxu Li, Maolei Fu, et~al.
\newblock Graphad: Interaction scene graph for end-to-end autonomous driving.
\newblock \emph{arXiv preprint arXiv:2403.19098}, 2024.

\bibitem[Jiang et~al.(2023)Jiang, Chen, Xu, Liao, Chen, Zhou, Zhang, Liu, Huang, and Wang]{E2E_VAD}
Bo~Jiang, Shaoyu Chen, Qing Xu, Bencheng Liao, Jiajie Chen, Helong Zhou, Qian Zhang, Wenyu Liu, Chang Huang, and Xinggang Wang.
\newblock Vad: Vectorized scene representation for efficient autonomous driving.
\newblock In \emph{Proceedings of the IEEE/CVF International Conference on Computer Vision}, pages 8340--8350, 2023.

\bibitem[Li et~al.(2024{\natexlab{b}})Li, Yu, Lan, Li, Kautz, Lu, and Alvarez]{E2E_BEVPlanner}
Zhiqi Li, Zhiding Yu, Shiyi Lan, Jiahan Li, Jan Kautz, Tong Lu, and Jose~M Alvarez.
\newblock Is ego status all you need for open-loop end-to-end autonomous driving?
\newblock In \emph{Proceedings of the IEEE/CVF Conference on Computer Vision and Pattern Recognition}, pages 14864--14873, 2024{\natexlab{b}}.

\bibitem[Weng et~al.(2024)Weng, Ivanovic, Wang, Wang, and Pavone]{E2E_paradrive}
Xinshuo Weng, Boris Ivanovic, Yan Wang, Yue Wang, and Marco Pavone.
\newblock Para-drive: Parallelized architecture for real-time autonomous driving.
\newblock In \emph{Proceedings of the IEEE/CVF Conference on Computer Vision and Pattern Recognition}, pages 15449--15458, 2024.

\bibitem[Wu et~al.(2022)Wu, Jia, Chen, Yan, Li, and Qiao]{E2E_TCP}
Penghao Wu, Xiaosong Jia, Li~Chen, Junchi Yan, Hongyang Li, and Yu~Qiao.
\newblock Trajectory-guided control prediction for end-to-end autonomous driving: A simple yet strong baseline.
\newblock \emph{Advances in Neural Information Processing Systems}, 35:\penalty0 6119--6132, 2022.

\bibitem[Zhang et~al.(2021)Zhang, Liniger, Dai, Yu, and Van~Gool]{E2E_roach}
Zhejun Zhang, Alexander Liniger, Dengxin Dai, Fisher Yu, and Luc Van~Gool.
\newblock End-to-end urban driving by imitating a reinforcement learning coach.
\newblock In \emph{Proceedings of the IEEE/CVF international conference on computer vision}, pages 15222--15232, 2021.

\bibitem[Jia et~al.(2023{\natexlab{a}})Jia, Wu, Chen, Xie, He, Yan, and Li]{E2E_thinktwice}
Xiaosong Jia, Penghao Wu, Li~Chen, Jiangwei Xie, Conghui He, Junchi Yan, and Hongyang Li.
\newblock Think twice before driving: Towards scalable decoders for end-to-end autonomous driving.
\newblock In \emph{Proceedings of the IEEE/CVF Conference on Computer Vision and Pattern Recognition}, pages 21983--21994, 2023{\natexlab{a}}.

\bibitem[Jia et~al.(2023{\natexlab{b}})Jia, Gao, Chen, Yan, Liu, and Li]{E2E_driveadapter}
Xiaosong Jia, Yulu Gao, Li~Chen, Junchi Yan, Patrick~Langechuan Liu, and Hongyang Li.
\newblock Driveadapter: Breaking the coupling barrier of perception and planning in end-to-end autonomous driving.
\newblock In \emph{Proceedings of the IEEE/CVF International Conference on Computer Vision}, pages 7953--7963, 2023{\natexlab{b}}.

\bibitem[Shao et~al.(2024)Shao, Hu, Wang, Song, Waslander, Liu, and Li]{E2E_lmdrive}
Hao Shao, Yuxuan Hu, Letian Wang, Guanglu Song, Steven~L Waslander, Yu~Liu, and Hongsheng Li.
\newblock Lmdrive: Closed-loop end-to-end driving with large language models.
\newblock In \emph{Proceedings of the IEEE/CVF Conference on Computer Vision and Pattern Recognition}, pages 15120--15130, 2024.

\bibitem[Sima et~al.(2024)Sima, Renz, Chitta, Chen, Zhang, Xie, Bei{\ss}wenger, Luo, Geiger, and Li]{E2E_drivelm}
Chonghao Sima, Katrin Renz, Kashyap Chitta, Li~Chen, Hanxue Zhang, Chengen Xie, Jens Bei{\ss}wenger, Ping Luo, Andreas Geiger, and Hongyang Li.
\newblock Drivelm: Driving with graph visual question answering.
\newblock In \emph{European Conference on Computer Vision}, pages 256--274. Springer, 2024.

\bibitem[Mao et~al.(2023)Mao, Qian, Ye, Zhao, and Wang]{E2E_Gpt-driver}
Jiageng Mao, Yuxi Qian, Junjie Ye, Hang Zhao, and Yue Wang.
\newblock Gpt-driver: Learning to drive with gpt.
\newblock \emph{arXiv preprint arXiv:2310.01415}, 2023.

\bibitem[Pan et~al.(2024)Pan, Yaman, Nesti, Mallik, Allievi, Velipasalar, and Ren]{E2E_vlp}
Chenbin Pan, Burhaneddin Yaman, Tommaso Nesti, Abhirup Mallik, Alessandro~G Allievi, Senem Velipasalar, and Liu Ren.
\newblock Vlp: Vision language planning for autonomous driving.
\newblock In \emph{Proceedings of the IEEE/CVF Conference on Computer Vision and Pattern Recognition}, pages 14760--14769, 2024.

\bibitem[Chen et~al.(2024{\natexlab{e}})Chen, Jiang, Gao, Liao, Xu, Zhang, Huang, Liu, and Wang]{E2E_vadv2}
Shaoyu Chen, Bo~Jiang, Hao Gao, Bencheng Liao, Qing Xu, Qian Zhang, Chang Huang, Wenyu Liu, and Xinggang Wang.
\newblock Vadv2: End-to-end vectorized autonomous driving via probabilistic planning.
\newblock \emph{arXiv preprint arXiv:2402.13243}, 2024{\natexlab{e}}.

\bibitem[Xu et~al.(2024)Xu, Hu, Zhang, Meyer, Mustikovela, Srinivasa, Wolff, and Huang]{E2E_Vlm-ad}
Yi~Xu, Yuxin Hu, Zaiwei Zhang, Gregory~P Meyer, Siva~Karthik Mustikovela, Siddhartha Srinivasa, Eric~M Wolff, and Xin Huang.
\newblock Vlm-ad: End-to-end autonomous driving through vision-language model supervision.
\newblock \emph{arXiv preprint arXiv:2412.14446}, 2024.

\bibitem[Madan et~al.(2021{\natexlab{a}})Madan, Henry, Dozier, Ho, Bhandari, Sasaki, Durand, Pfister, and Boix]{madan2021and}
Spandan Madan, Timothy Henry, Jamell Dozier, Helen Ho, Nishchal Bhandari, Tomotake Sasaki, Fr{\'e}do Durand, Hanspeter Pfister, and Xavier Boix.
\newblock When and how cnns generalize to outof-distribution category-viewpoint combinations.
\newblock \emph{arXiv preprint arXiv:2007.08032}, 2021{\natexlab{a}}.

\bibitem[Madan et~al.(2021{\natexlab{b}})Madan, Sasaki, Li, Boix, and Pfister]{madan2021small}
Spandan Madan, Tomotake Sasaki, Tzu-Mao Li, Xavier Boix, and Hanspeter Pfister.
\newblock Small in-distribution changes in 3d perspective and lighting fool both cnns and transformers.
\newblock \emph{arXiv preprint arXiv:2106.16198}, 3, 2021{\natexlab{b}}.

\bibitem[Coors et~al.(2019)Coors, Condurache, and Geiger]{coors2019nova}
Benjamin Coors, Alexandru~Paul Condurache, and Andreas Geiger.
\newblock Nova: Learning to see in novel viewpoints and domains.
\newblock In \emph{2019 International Conference on 3D Vision (3DV)}, pages 116--125. IEEE, 2019.

\bibitem[Do et~al.(2020)Do, Vuong, Roumeliotis, and Park]{do2020surface}
Tien Do, Khiem Vuong, Stergios~I Roumeliotis, and Hyun~Soo Park.
\newblock Surface normal estimation of tilted images via spatial rectifier.
\newblock In \emph{Computer Vision--ECCV 2020: 16th European Conference, Glasgow, UK, August 23--28, 2020, Proceedings, Part IV 16}, pages 265--280. Springer, 2020.

\bibitem[Gao et~al.(2024)Gao, Chen, Li, Hong, Li, and Xu]{gao2024magicdrive3d}
Ruiyuan Gao, Kai Chen, Zhihao Li, Lanqing Hong, Zhenguo Li, and Qiang Xu.
\newblock Magicdrive3d: Controllable 3d generation for any-view rendering in street scenes.
\newblock \emph{arXiv preprint arXiv:2405.14475}, 2024.

\bibitem[Chang et~al.(2024)Chang, Lee, Kim, Kim, Lee, Ji, Jang, and Kim]{chang2024unified}
Gyusam Chang, Jiwon Lee, Donghyun Kim, Jinkyu Kim, Dongwook Lee, Daehyun Ji, Sujin Jang, and Sangpil Kim.
\newblock Unified domain generalization and adaptation for multi-view 3d object detection.
\newblock \emph{Advances in Neural Information Processing Systems}, 37:\penalty0 58498--58524, 2024.

\bibitem[Yan et~al.(2024)Yan, Lin, Zhou, Wang, Sun, Zhan, Lang, Zhou, and Peng]{yan2024street}
Yunzhi Yan, Haotong Lin, Chenxu Zhou, Weijie Wang, Haiyang Sun, Kun Zhan, Xianpeng Lang, Xiaowei Zhou, and Sida Peng.
\newblock Street gaussians: Modeling dynamic urban scenes with gaussian splatting.
\newblock In \emph{European Conference on Computer Vision}, pages 156--173. Springer, 2024.

\bibitem[Wu et~al.(2024)Wu, Yi, Fang, Xie, Zhang, Wei, Liu, Tian, and Wang]{wu20244d}
Guanjun Wu, Taoran Yi, Jiemin Fang, Lingxi Xie, Xiaopeng Zhang, Wei Wei, Wenyu Liu, Qi~Tian, and Xinggang Wang.
\newblock 4d gaussian splatting for real-time dynamic scene rendering.
\newblock In \emph{Proceedings of the IEEE/CVF conference on computer vision and pattern recognition}, pages 20310--20320, 2024.

\bibitem[Lin et~al.(2024)Lin, Li, Tang, Liu, Liu, Liu, Lu, Wu, Xu, Yan, et~al.]{lin2024vastgaussian}
Jiaqi Lin, Zhihao Li, Xiao Tang, Jianzhuang Liu, Shiyong Liu, Jiayue Liu, Yangdi Lu, Xiaofei Wu, Songcen Xu, Youliang Yan, et~al.
\newblock Vastgaussian: Vast 3d gaussians for large scene reconstruction.
\newblock In \emph{Proceedings of the IEEE/CVF Conference on Computer Vision and Pattern Recognition}, pages 5166--5175, 2024.

\bibitem[Turki et~al.(2024)Turki, Agrawal, Bul{\`o}, Porzi, Kontschieder, Ramanan, Zollh{\"o}fer, and Richardt]{turki2024hybridnerf}
Haithem Turki, Vasu Agrawal, Samuel~Rota Bul{\`o}, Lorenzo Porzi, Peter Kontschieder, Deva Ramanan, Michael Zollh{\"o}fer, and Christian Richardt.
\newblock Hybridnerf: Efficient neural rendering via adaptive volumetric surfaces.
\newblock In \emph{Proceedings of the IEEE/CVF Conference on Computer Vision and Pattern Recognition}, pages 19647--19656, 2024.

\bibitem[Kerbl et~al.(2023)Kerbl, Kopanas, Leimk{\"u}hler, and Drettakis]{kerbl20233d}
Bernhard Kerbl, Georgios Kopanas, Thomas Leimk{\"u}hler, and George Drettakis.
\newblock 3d gaussian splatting for real-time radiance field rendering.
\newblock \emph{ACM Trans. Graph.}, 42\penalty0 (4):\penalty0 139--1, 2023.

\bibitem[Szymanowicz et~al.(2024)Szymanowicz, Rupprecht, and Vedaldi]{szymanowicz2024splatter}
Stanislaw Szymanowicz, Chrisitian Rupprecht, and Andrea Vedaldi.
\newblock Splatter image: Ultra-fast single-view 3d reconstruction.
\newblock In \emph{Proceedings of the IEEE/CVF conference on computer vision and pattern recognition}, pages 10208--10217, 2024.

\bibitem[Zheng et~al.(2024{\natexlab{c}})Zheng, Zhou, Shao, Liu, Zhang, Nie, and Liu]{zheng2024gps}
Shunyuan Zheng, Boyao Zhou, Ruizhi Shao, Boning Liu, Shengping Zhang, Liqiang Nie, and Yebin Liu.
\newblock Gps-gaussian: Generalizable pixel-wise 3d gaussian splatting for real-time human novel view synthesis.
\newblock In \emph{Proceedings of the IEEE/CVF conference on computer vision and pattern recognition}, pages 19680--19690, 2024{\natexlab{c}}.

\bibitem[He et~al.(2016)He, Zhang, Ren, and Sun]{resnet}
Kaiming He, Xiangyu Zhang, Shaoqing Ren, and Jian Sun.
\newblock Deep residual learning for image recognition.
\newblock In \emph{Proceedings of the IEEE conference on computer vision and pattern recognition}, pages 770--778, 2016.

\bibitem[Sun et~al.(2024)Sun, Lin, Shi, Zhang, Wu, and Zheng]{E2E_Sparsedrive}
Wenchao Sun, Xuewu Lin, Yining Shi, Chuang Zhang, Haoran Wu, and Sifa Zheng.
\newblock Sparsedrive: End-to-end autonomous driving via sparse scene representation.
\newblock \emph{arXiv preprint arXiv:2405.19620}, 2024.

\bibitem[Jia et~al.(2025)Jia, You, Zhang, and Yan]{E2E_drivetransformer}
Xiaosong Jia, Junqi You, Zhiyuan Zhang, and Junchi Yan.
\newblock Drivetransformer: Unified transformer for scalable end-to-end autonomous driving.
\newblock \emph{arXiv preprint arXiv:2503.07656}, 2025.

\bibitem[Lin et~al.(2022)Lin, Lin, Pei, Huang, and Su]{lin2022sparse4d}
Xuewu Lin, Tianwei Lin, Zixiang Pei, Lichao Huang, and Zhizhong Su.
\newblock Sparse4d: Multi-view 3d object detection with sparse spatial-temporal fusion.
\newblock \emph{arXiv preprint arXiv:2211.10581}, 2022.

\bibitem[Lin et~al.(2023)Lin, Pei, Lin, Huang, and Su]{lin2023sparse4d}
Xuewu Lin, Zixiang Pei, Tianwei Lin, Lichao Huang, and Zhizhong Su.
\newblock Sparse4d v3: Advancing end-to-end 3d detection and tracking.
\newblock \emph{arXiv preprint arXiv:2311.11722}, 2023.

\bibitem[Vaswani et~al.(2017)Vaswani, Shazeer, Parmar, Uszkoreit, Jones, Gomez, Kaiser, and Polosukhin]{vaswani2017attention}
Ashish Vaswani, Noam Shazeer, Niki Parmar, Jakob Uszkoreit, Llion Jones, Aidan~N Gomez, {\L}ukasz Kaiser, and Illia Polosukhin.
\newblock Attention is all you need.
\newblock \emph{Advances in neural information processing systems}, 30, 2017.

\bibitem[Wang et~al.(2022)Wang, Guizilini, Zhang, Wang, Zhao, and Solomon]{wang2022detr3d}
Yue Wang, Vitor~Campagnolo Guizilini, Tianyuan Zhang, Yilun Wang, Hang Zhao, and Justin Solomon.
\newblock Detr3d: 3d object detection from multi-view images via 3d-to-2d queries.
\newblock In \emph{Conference on Robot Learning}, pages 180--191. PMLR, 2022.

\bibitem[Liang et~al.(2020)Liang, Yang, Hu, Chen, Liao, Feng, and Urtasun]{liang2020learning}
Ming Liang, Bin Yang, Rui Hu, Yun Chen, Renjie Liao, Song Feng, and Raquel Urtasun.
\newblock Learning lane graph representations for motion forecasting.
\newblock In \emph{Computer Vision--ECCV 2020: 16th European Conference, Glasgow, UK, August 23--28, 2020, Proceedings, Part II 16}, pages 541--556. Springer, 2020.

\bibitem[Lin et~al.(2017)Lin, Goyal, Girshick, He, and Doll{\'a}r]{lin2017focal}
Tsung-Yi Lin, Priya Goyal, Ross Girshick, Kaiming He, and Piotr Doll{\'a}r.
\newblock Focal loss for dense object detection.
\newblock In \emph{Proceedings of the IEEE international conference on computer vision}, pages 2980--2988, 2017.

\bibitem[Zhang et~al.(2018)Zhang, Isola, Efros, Shechtman, and Wang]{zhang2018unreasonable}
Richard Zhang, Phillip Isola, Alexei~A Efros, Eli Shechtman, and Oliver Wang.
\newblock The unreasonable effectiveness of deep features as a perceptual metric.
\newblock In \emph{Proceedings of the IEEE conference on computer vision and pattern recognition}, pages 586--595, 2018.

\bibitem[Hu et~al.(2022)Hu, Chen, Wu, Li, Yan, and Tao]{E2E_ST-P3}
Shengchao Hu, Li~Chen, Penghao Wu, Hongyang Li, Junchi Yan, and Dacheng Tao.
\newblock St-p3: End-to-end vision-based autonomous driving via spatial-temporal feature learning.
\newblock In \emph{European Conference on Computer Vision}, pages 533--549. Springer, 2022.

\bibitem[Zhai et~al.(2023)Zhai, Feng, Du, Mao, Liu, Tan, Zhang, Ye, and Wang]{E2E_ADMLP_archeive}
Jiang-Tian Zhai, Ze~Feng, Jinhao Du, Yongqiang Mao, Jiang-Jiang Liu, Zichang Tan, Yifu Zhang, Xiaoqing Ye, and Jingdong Wang.
\newblock Rethinking the open-loop evaluation of end-to-end autonomous driving in nuscenes.
\newblock \emph{arXiv preprint arXiv:2305.10430}, 2023.

\bibitem[Caesar et~al.(2020)Caesar, Bankiti, Lang, Vora, Liong, Xu, Krishnan, Pan, Baldan, and Beijbom]{Benchmarks_NuScenes}
Holger Caesar, Varun Bankiti, Alex~H Lang, Sourabh Vora, Venice~Erin Liong, Qiang Xu, Anush Krishnan, Yu~Pan, Giancarlo Baldan, and Oscar Beijbom.
\newblock nuscenes: A multimodal dataset for autonomous driving.
\newblock In \emph{Proceedings of the IEEE/CVF conference on computer vision and pattern recognition}, pages 11621--11631, 2020.

\bibitem[Chen et~al.(2024{\natexlab{f}})Chen, Yang, Huang, de~Lutio, Esturo, Ivanovic, Litany, Gojcic, Fidler, Pavone, et~al.]{chen2024omnire}
Ziyu Chen, Jiawei Yang, Jiahui Huang, Riccardo de~Lutio, Janick~Martinez Esturo, Boris Ivanovic, Or~Litany, Zan Gojcic, Sanja Fidler, Marco Pavone, et~al.
\newblock Omnire: Omni urban scene reconstruction.
\newblock \emph{arXiv preprint arXiv:2408.16760}, 2024{\natexlab{f}}.

\bibitem[Dosovitskiy et~al.(2017)Dosovitskiy, Ros, Codevilla, Lopez, and Koltun]{Benchmarks_Carla}
Alexey Dosovitskiy, German Ros, Felipe Codevilla, Antonio Lopez, and Vladlen Koltun.
\newblock Carla: An open urban driving simulator.
\newblock In \emph{Conference on robot learning}, pages 1--16. PMLR, 2017.

\bibitem[Prakash et~al.(2021)Prakash, Chitta, and Geiger]{E2E_Tranfuser}
Aditya Prakash, Kashyap Chitta, and Andreas Geiger.
\newblock Multi-modal fusion transformer for end-to-end autonomous driving.
\newblock In \emph{Proceedings of the IEEE/CVF conference on computer vision and pattern recognition}, pages 7077--7087, 2021.

\end{thebibliography}

\end{document}